\documentclass[a4paper,fleqn,10.5pt]{cas-dc}
\usepackage[utf8]{inputenc}
\usepackage[numbers,sort&compress]{natbib}
\usepackage[dvipsnames]{xcolor}
\usepackage[labelfont=bf]{caption}
\usepackage{subcaption}
\usepackage{epstopdf}
\usepackage{threeparttable}
\usepackage{etoolbox}
\usepackage{hyperref}
\usepackage{graphicx}
\usepackage{amsmath}
\usepackage{amssymb}
\usepackage{booktabs}
\usepackage{multirow}
\usepackage{float}
\usepackage{placeins}
\usepackage{needspace}
\usepackage{tabularx}
\usepackage{algorithm}
\usepackage{algpseudocode}

\def\tsc#1{\csdef{#1}{\textsc{\lowercase{#1}}\xspace}}
\tsc{WGM}\tsc{QE}\tsc{EP}\tsc{PMS}\tsc{BEC}

\begin{document}
\let\WriteBookmarks\relax

\renewcommand{\topfraction}{0.97}
\renewcommand{\bottomfraction}{0.97}
\renewcommand{\textfraction}{0.03}
\renewcommand{\floatpagefraction}{0.94}
\setcounter{topnumber}{6}
\setcounter{bottomnumber}{4}
\setcounter{totalnumber}{8}
\setlength{\textfloatsep}{4pt plus 1pt minus 1pt}
\setlength{\floatsep}{3pt plus 1pt minus 1pt}
\setlength{\intextsep}{3pt plus 1pt minus 1pt}
\setlength{\abovecaptionskip}{1pt}
\setlength{\belowcaptionskip}{0pt}
\setlength{\dbltextfloatsep}{4pt plus 1pt minus 1pt}
\setlength{\dblfloatsep}{3pt plus 1pt minus 1pt}

\shorttitle{Physics-Guided ML Extrapolation Framework Using a Classical Transient Diffusion Benchmark}
\shortauthors{A.~Yadav~et~al.}

\title[mode=title]{Development and Validation of a Physics-Guided Machine Learning Extrapolation Framework Using a Classical Transient Diffusion Benchmark}

\author[1]{Ashutosh Yadav}[orcid=0000-0000-0000-0000]
\author[1]{Alok Dubey}[orcid=0009-0004-8021-6262]
\author[1]{Prodyut Ranjan Chakraborty}[orcid=0000-0002-9456-6747]
\ead{pchakraborty@iitj.ac.in}
\author[2]{Harshal Deepak Akolekar}[orcid=0000-0000-0002-3178-2987]
\ead{harshal.akolekar@iitj.ac.in}
\cormark[1]

\address[1]{Department of Mechanical Engineering,
            Indian Institute of Technology Jodhpur,
            Jodhpur 342030, India}
            \address[2]{Department of Aerospace Engineering,
            Indian Institute of Technology Jodhpur,
            Jodhpur 342030, India}
\cortext[cor1]{Corresponding author}

\begin{abstract}
Machine learning models employed in engineering applications are typically trained using data confined to a limited operating range; however, reliable predictions are often required beyond this domain. Consequently, the primary challenge lies not in interpolation but in extrapolation. Rigorous validation of extrapolation performance on real engineering systems is often hindered by the scarcity of data outside the training range. To address this limitation, a novel extrapolation framework is integrated with established machine learning architectures to enable accurate and physically consistent predictions beyond the training domain. To establish the proposed extrapolation framework, the performance of two physics-guided machine learning architectures, namely a Bidirectional Long Short-Term Memory (BiLSTM) network and a Physics-Informed Neural Network (PINN), is systematically evaluated. To validate the proposed extrapolation framework, a classical one-dimensional transient diffusion problem is adopted as a benchmark owing to the availability of an exact analytical solution. The analytical solution provides unlimited, reliable data across the entire spatio-temporal domain, enabling rigorous quantitative validation of extrapolated predictions. The benchmark is particularly challenging because the solution evolves from an initial singularity through a strongly nonlinear transient regime before approaching a steady-state linear profile. When training data are confined to an intermediate portion of this evolution, backward extrapolation toward the singularity becomes especially demanding. To improve predictive reliability in this regime, physics-guided coordinate transformations, boundary-aware learning strategies, and stability-enhancing temporal marching are incorporated into the learning framework. Extrapolation performance is evaluated using a train–predict–validate–extend strategy, wherein validated predictions are recursively incorporated into the training set to progressively extend the prediction horizon. The results demonstrate accurate and physically consistent predictions beyond the training domain, indicating the potential of the proposed framework for a wide range of engineering applications characterized by limited data availability.

\end{abstract}


\maketitle

\begin{table*}[!t]
\centering
\caption*{\textbf{Nomenclature}}
\small
\setlength{\tabcolsep}{4pt}
\renewcommand{\arraystretch}{1.0}
\begin{tabular}{@{}p{0.485\textwidth}@{\hspace{0.02\textwidth}}p{0.475\textwidth}@{}}
\toprule
\textbf{Roman symbols} & \textbf{Greek symbols} \\
\midrule
\begin{tabularx}{\linewidth}{@{}l@{\hspace{6pt}}>{\raggedright\arraybackslash}X@{}}
$A$            & Amplitude factor \\
$A_n$          & Modal amplitude \\
$C_n$          & Fourier coefficient \\
$Fo$           & Fourier number \\
$g(x^*)$       & Steady-state component (hard-BC ansatz) \\
$h(x^*)$       & Blending function (hard-BC ansatz) \\
$K$            & Sliding-window width \\
$L$            & Rod length \\
$\mathcal{N}$  & Network output (hard-BC ansatz) \\
$n$            & Mode index \\
$n_{\mathrm{eff}}$ & Effective mode count \\
$N_{Fo}$       & Number of Fourier-number samples \\
$N_{IC}$       & Number of initial-condition points \\
$N_r$          & Number of PDE collocation points \\
$N_x$          & Number of spatial grid points \\
$\mathcal{R}$  & PDE residual \\
$s$            & Root-Fourier coordinate \\
$t$            & Time \\
$T$            & Temperature \\
$T_i$          & Initial temperature \\
$T_L$          & Left-boundary temperature \\
$T_R$          & Right-boundary temperature \\
$w(x^*)$       & Boundary weight function \\
$x$            & Spatial coordinate \\
$x^*$          & Dimensionless coordinate \\
\end{tabularx}
&
\begin{tabularx}{\linewidth}{@{}l@{\hspace{6pt}}>{\raggedright\arraybackslash}X@{}}
$\alpha$         & Thermal diffusivity \\
$\delta^*$       & Dimensionless penetration depth \\
$\varepsilon$    & Log-time regularising constant \\
$\theta$         & Dimensionless temperature \\
$\theta_\infty$  & Steady-state dimensionless temperature \\
$\lambda$        & Growth factor / loss-weighting coefficient \\
$\rho$           & Relaxed convergence factor \\
$\sigma(\cdot)$  & Sigmoid activation function \\
$\sigma_w$       & Gaussian width parameter \\
$\tau$           & Log-time coordinate \\
$\omega$         & Under-relaxation factor \\
\end{tabularx} \\[7pt]
\multicolumn{2}{@{}l@{}}{\textbf{Subscripts and superscripts}} \\[2pt]
\multicolumn{2}{@{}l@{}}{
\begin{tabularx}{\textwidth}{@{}l@{\hspace{6pt}}>{\raggedright\arraybackslash}p{0.235\textwidth}@{\hspace{12pt}}l@{\hspace{6pt}}>{\raggedright\arraybackslash}X@{}}
$BW$   & Boundary-weighted (loss) & $n$              & Mode index \\
data   & Data-fidelity (loss)     & $PDE$            & Governing-equation (residual/loss) \\
eff    & Effective quantity       & $R$              & Right boundary \\
$i$    & Initial condition        & $*$              & Dimensionless quantity \\
$IC$   & Initial-condition (loss) & $\infty$         & Steady-state value \\
$k$    & Time-step index          & $\hat{(\cdot)}$  & Predicted quantity \\
$L$    & Left boundary            &                  & \\
\end{tabularx}} \\
\bottomrule
\end{tabular}
\end{table*}

\section{Introduction}
\label{sec:intro}

Machine learning (ML) has become an increasingly important tool for modeling complex engineering systems using data derived from experiments, numerical simulations, and operational measurements. However, training datasets are often confined to a limited range of operating conditions due to practical constraints on data generation. In contrast, engineering design and decision-making frequently require predictions beyond the available data range, where additional experiments or high-fidelity simulations may be prohibitively expensive or infeasible. Consequently, the primary challenge in engineering machine learning is not interpolation within the training domain, but reliable extrapolation beyond it. Yet, extrapolation presents a fundamental dilemma: how can the credibility of a prediction be established when the reference data needed for validation are inherently scarce or unavailable outside the training domain? Overcoming this limitation represents one of the most important unresolved challenges in data-driven engineering.

Rigorous assessment of ML-assisted extrapolative prediction requires a benchmark problem with three essential characteristics. First, an analytical or otherwise exact reference solution must be available throughout the entire spatio-temporal domain, including regions well beyond the training window, so that unlimited reference data can be generated at any desired spatio-temporal location. Second, extrapolated predictions must be quantitatively validated through direct comparison with the reference solution in the extrapolation region. Third, the benchmark must exhibit strongly nonlinear behavior, ensuring that successful extrapolation cannot be attributed merely to fitting a smooth function. A benchmark lacking these characteristics provides little evidence of true extrapolative capability.

A one-dimensional (1-D) transient heat-conduction problem with Dirichlet boundary conditions satisfies all the aforementioned requirements and, therefore, serves as an ideal benchmark problem. First, it admits an analytical series solution over the entire spatio-temporal domain. Furthermore, the non-dimensionalization of the problem in terms of the non-dimensional time (Fourier number) $F_o \in \langle 0,\infty \rangle$, non-dimensional spatial coordinate $x^{*} \in \langle 0,1 \rangle$, and non-dimensional temperature $\theta \in \langle 0,1 \rangle$ renders the solution applicable to the entire class of such problems, irrespective of the specific values of thermal diffusivity, characteristic length scale, time duration, and the prescribed initial and boundary conditions. Consequently, the analytical solution represents a generalized solution for this class of one-dimensional transient heat-conduction problems.

Second, the availability of the analytical solution over the complete spatio-temporal domain ($F_o \in \langle 0,\infty \rangle$, $x^{*} \in \langle 0,1 \rangle$) enables rigorous validation of ML-derived interpolated as well as extrapolated solutions at any point within the domain. This comprehensive coverage provides a robust framework for assessing the predictive capability and generalization performance of machine-learning models.

Third, the solution undergoes a highly nonlinear transient evolution, beginning with singularities at the boundaries at $F_o=0$ and gradually approaching the steady-state linear solution of the form $\theta (x^*)=\theta (x^* =0) \pm x^*$ as $F_o \to \infty$. This continuous transition from a strongly nonlinear transient regime to a linear steady-state solution makes this class of problems particularly suitable for evaluating the capability of ML models to capture complex nonlinear behavior. As expected, the degree of nonlinearity is highest in the vicinity of $F_o \to 0$ and progressively diminishes with increasing Fourier number, eventually vanishing as the solution approaches the steady state. 

Although the primary objective of the present work is to develop and assess an ML-based extrapolation strategy with broad applicability to diverse classes of data-driven problems, the benchmark problem considered herein is itself of considerable practical significance. Accurate prediction of the transient temperature distribution within a solid is fundamental to numerous engineering applications. For example, in multi-core processors, transient thermal analysis enables the prediction of localized hot-spots before they compromise device reliability or performance~\cite{Lu2024thermalmap,Niederer2021}. In metal additive manufacturing, it facilitates precise control of melt-pool thermal history, which directly influences microstructural evolution and part quality~\cite{Yan2018,Oommen2022}. Likewise, in aerospace structures, transient heat-conduction analysis plays a critical role in evaluating and certifying structural integrity under severe aerodynamic heating conditions~\cite{Willard2022}. Consequently, the selected benchmark not only provides a rigorous framework for evaluating ML-based extrapolation techniques but also represents a problem of substantial engineering relevance.

Although physics-based numerical methods, such as the finite-element and finite-difference methods, are capable of accurately predicting transient thermal fields, their computational cost often renders them unsuitable for applications requiring real-time control, inverse parameter estimation, uncertainty quantification, or large-scale design optimization~\cite{Kapteyn2021,EbbsPicken2024}. This computational bottleneck has motivated the increasing adoption of machine-learning surrogates that can approximate the underlying physics with orders-of-magnitude faster inference times~\cite{Karniadakis2021,Willard2022}. However, the practical utility of such surrogates extends well beyond interpolation, as many engineering applications inevitably require predictions in previously unexplored regions of the parameter or temporal domain. Consequently, the computational advantages offered by ML models are meaningful only if their extrapolated predictions remain accurate and physically reliable. Establishing such confidence through rigorous validation, therefore, constitutes the central motivation of the present work.

What makes the heat-conduction testbed particularly demanding is its behavior at small Fourier numbers, where the temperature field evolves rapidly and forward and backward extrapolation become fundamentally asymmetric.  Forward prediction is relatively well-behaved because thermal diffusion progressively smooths the temperature field, causing small prediction errors to decay over time.  Backward prediction is considerably more challenging because it requires reconstructing an earlier, sharper temperature field from a later, more diffused state. This effectively reverses the smoothing process and amplifies small errors, making the backward heat equation a classical ill-posed problem~\cite{Hadamard1902,Beck1985}. Consequently, accurate extrapolation in both temporal directions provides a stringent test of a surrogate's ability to capture the underlying diffusion dynamics rather than merely interpolate or memorize the training data. This makes the benchmark particularly valuable for assessing whether a surrogate has learned the governing physics and can reliably generalize beyond the conditions encountered during training.

Long short-term memory (LSTM) networks~\cite{Hochreiter1997} and their bidirectional extension, BiLSTM~\cite{Schuster1997}, constitute natural candidates for data-driven temporal extrapolation because of their ability to capture long-range temporal dependencies in sequential data. These architectures have demonstrated remarkable success in learning the temporal evolution of temperature and flow fields, thereby providing computationally efficient surrogate models for applications including electronic cooling~\cite{Xiang2025bilstm}, fluid dynamics~\cite{Vlachas2018}, turbulence modeling~\cite{Srinivasan2019}, and structural dynamics~\cite{Zhang2020lstm}. Furthermore, Graves \emph{et al.}~\cite{Graves2013} showed that incorporating temporal information from both forward and backward directions significantly enhances predictive accuracy in sequence-learning tasks, thereby motivating the use of BiLSTM architectures for complex transient phenomena.

Despite these successes, existing studies almost exclusively develop and evaluate their models within the same finite temporal domain used for training. Consequently, the reported predictive performance primarily reflects interpolation or short-horizon forecasting rather than true extrapolation. More importantly, these investigations do not examine whether the learned models can recursively propagate predictions into regions where the underlying physical behavior undergoes qualitative changes, nor do they validate such predictions against analytical or otherwise exact solutions far beyond the training window. Addressing this critical gap constitutes the central objective of the present work, which systematically investigates the extrapolation capability of LSTM-based models under rigorously verifiable conditions.

Physics-informed neural networks (PINNs)~\cite{Raissi2019,Karniadakis2021} represent a second promising framework for data-driven extrapolation, adopting a fundamentally different learning paradigm from conventional neural networks. Rather than relying exclusively on labeled data, PINNs incorporates the governing partial differential equations and associated initial and boundary conditions directly into the training objective, thereby enforcing physical consistency throughout the learning process. This formulation enables PINNs to achieve accurate predictions even in regions where labeled data are sparse or unavailable, as demonstrated by Cai \emph{et al.}~\cite{Cai2021} for both forward and inverse heat-transfer problems involving complex geometries, with subsequent extensions to transient and coupled multi-physics systems~\cite{Mattey2022novel,Zobeiry2021}. Nevertheless, subsequent studies have shown that training PINNs can be challenging when the components of the composite loss function are poorly balanced, leading to slow convergence or suboptimal solutions~\cite{Wang2021ntk}. To alleviate these issues, several strategies have been proposed, including curriculum-based expansion of the training domain~\cite{Mattey2022novel,bengio2009} and adaptive loss-weighting techniques~\cite{wight2021,mcclenny2023}. Despite these advances, existing studies have largely focused on improving training efficiency and solution accuracy within the prescribed computational domain. The capability of PINNs to extrapolate beyond the training domain has received comparatively little attention. In particular, a systematic evaluation of extrapolated PINNs predictions against an analytical or otherwise exact reference solution remains absent from the literature. Addressing this gap constitutes one of the principal objectives of the present work.

As an initial assessment, a Bi-LSTM and a PINN~\cite{Raissi2019} are trained using the transient one-dimensional heat-conduction benchmark with Dirichlet boundary conditions, where the training dataset comprises solutions corresponding to a prescribed range of Fourier numbers. The objective is not to demonstrate their ability to reproduce the thermal field within the training domain, since both architectures achieve satisfactory in-domain accuracy, but rather to investigate whether they learn the underlying physical dynamics sufficiently well to enable reliable extrapolation beyond the training data. This distinction becomes particularly evident during extrapolation. For both models, the most challenging scenario corresponds to backward extrapolation in the temporal domain, where prediction errors increase substantially. The degradation in performance is especially pronounced in the vicinity of $F_o \rightarrow 0$, where the transient diffusion process is characterized by strong nonlinear behavior and increased stiffness, such that small variations in the Fourier number ($F_o$) produce disproportionately large changes in the temperature field. These observations demonstrate that high predictive accuracy within the training domain does not necessarily translate into reliable physical extrapolation~\cite{Xu2021}. Although conventional sequence-learning and physics-informed architectures can reproduce the governing behavior in regions covered by the training data, they fail to consistently capture the underlying dynamics needed for robust prediction in previously unseen regimes~\cite{Bhatt2023,Kim2021DPM}. This limitation highlights a fundamental challenge for existing ML architectures when the primary objective extends beyond interpolation to physics-consistent extrapolation outside the training domain.

A closer examination suggests that these shortcomings originate from a mismatch between the mathematical representations adopted by the learning algorithms and the intrinsic structure of the underlying physics. In a conventional recurrent neural network, the Fourier number ($F_o$) is treated as an ordinary linear input variable, whereas the characteristic thermal penetration depth evolves proportionally to $\sqrt{F_o}$~\cite{Carslaw1959,Incropera2007}, as derived in Section~\ref{sec:governing}. This inconsistency becomes particularly significant during the early stages of diffusion, where the rapidly developing thermal boundary layer causes small prediction errors to accumulate and propagate through successive recurrent evaluations~\cite{Brandstetter2022}. PINNs encounter a different, yet fundamentally related, limitation. Because the boundary conditions are imposed indirectly through the loss function rather than being satisfied exactly by construction, even small boundary residuals can propagate into the solution interior through the governing partial differential equation (PDE)~\cite{Raissi2019}. Furthermore, the presence of the $F_o^{-3/2}$ gradient singularity renders the early-time regime particularly challenging for gradient-based optimization, often leading to poor convergence and degraded predictive accuracy~\cite{mcclenny2023,Krishnapriyan2021}. These observations indicate that the principal limitation is not the expressive capacity of the neural networks themselves, but rather the incompatibility between their mathematical representations and the governing physical structure of the problem. Consequently, no first-principle justification exists for expecting conventional formulations to naturally overcome these deficiencies~\cite{Xu2021}. The central question addressed in the present work is therefore whether physically motivated modifications to the BiLSTM and PINN frameworks can better align the learning process with the underlying diffusion physics, thereby enabling accurate, physics-consistent extrapolation across the entire computational domain when validated against the exact analytical solution.

Establishing the reliability of ML-based extrapolation requires more than the demonstration of a single successful prediction; rather, it demands a systematic, rigorous, and verifiable validation methodology, which constitutes one of the principal contributions of the present work. The availability of an exact analytical solution over the entire Fourier-number domain ($F_o$) provides a unique opportunity to develop such a framework. To this end, a finite interval of $F_o$ is selected and partitioned into three contiguous subdomains. The central subdomain is designated as the available-data region and serves as the initial training dataset, whereas the two adjacent subdomains are reserved exclusively for extrapolation training and validation. Beginning from the boundary of the available-data region, the ML model predicts the solution at the immediately adjacent Fourier number located within the extrapolation training domain. This prediction is then compared with the corresponding analytical solution, and an iterative error-minimization procedure is employed to systematically adjust the extrapolation parameters, including the data-fitting coefficients, data density, correction factors, and related quantities, until the prediction converges to the desired level of accuracy. Once validated, the extrapolated value replaces the corresponding analytical value in the available-data region and is subsequently used as an input for predicting the next point in the extrapolation sequence. Repeating this procedure progressively advances the extrapolation front while maintaining an exact analytical verification at every prediction step. In addition to producing accurate extrapolated solutions, this sequential framework also reveals the evolution of the extrapolation parameters, such as the data-fitting coefficients, data density, and correction factors, throughout the extrapolation process. These evolving parameter trends constitute a physics-guided extrapolation strategy that can subsequently be transferred beyond the training domain. In practical applications, where analytical solutions or measured data outside the training region are unavailable or sparse, the extrapolation procedure becomes fully deterministic, relying exclusively on the parameter-evolution patterns established during the validated training phase.

Overall, the present work proposes two complementary physics-guided, ML-based extrapolation frameworks whose applicability extends well beyond the one-dimensional transient heat-conduction benchmark considered herein. Although the benchmark problem serves as a rigorous and analytically verifiable testbed for the development and validation of the proposed methodologies, the underlying principles are sufficiently general to be explored for a broad range of data-driven problems in science and engineering. The BiLSTM-based framework is particularly attractive for applications where only limited prior physical knowledge is available, as it relies primarily on learning the temporal evolution directly from the data while incorporating minimal physics-guided modifications. In contrast, the PINN-based framework leverages the availability of governing equations and associated physical constraints to produce more accurate, physically consistent extrapolative predictions. In particular, its strong performance in backward extrapolation is consistent with PINNs' inherent ability to address inverse problems, such as reconstructing unknown initial conditions from subsequent observations. Together, these two frameworks provide complementary extrapolation strategies, enabling practitioners to select the most appropriate approach according to the availability of prior physical knowledge and the characteristics of the problem under consideration. The remainder of this paper presents the formulation, implementation, and validation of these two extrapolation frameworks and demonstrates their performance relative to exact analytical solutions over the entire computational domain.

\section{Methodology}\label{sec:governing}

\subsection{Governing Equations}
\label{subsec:problem}

\begin{figure}
\centering
\includegraphics[width=.98\columnwidth]{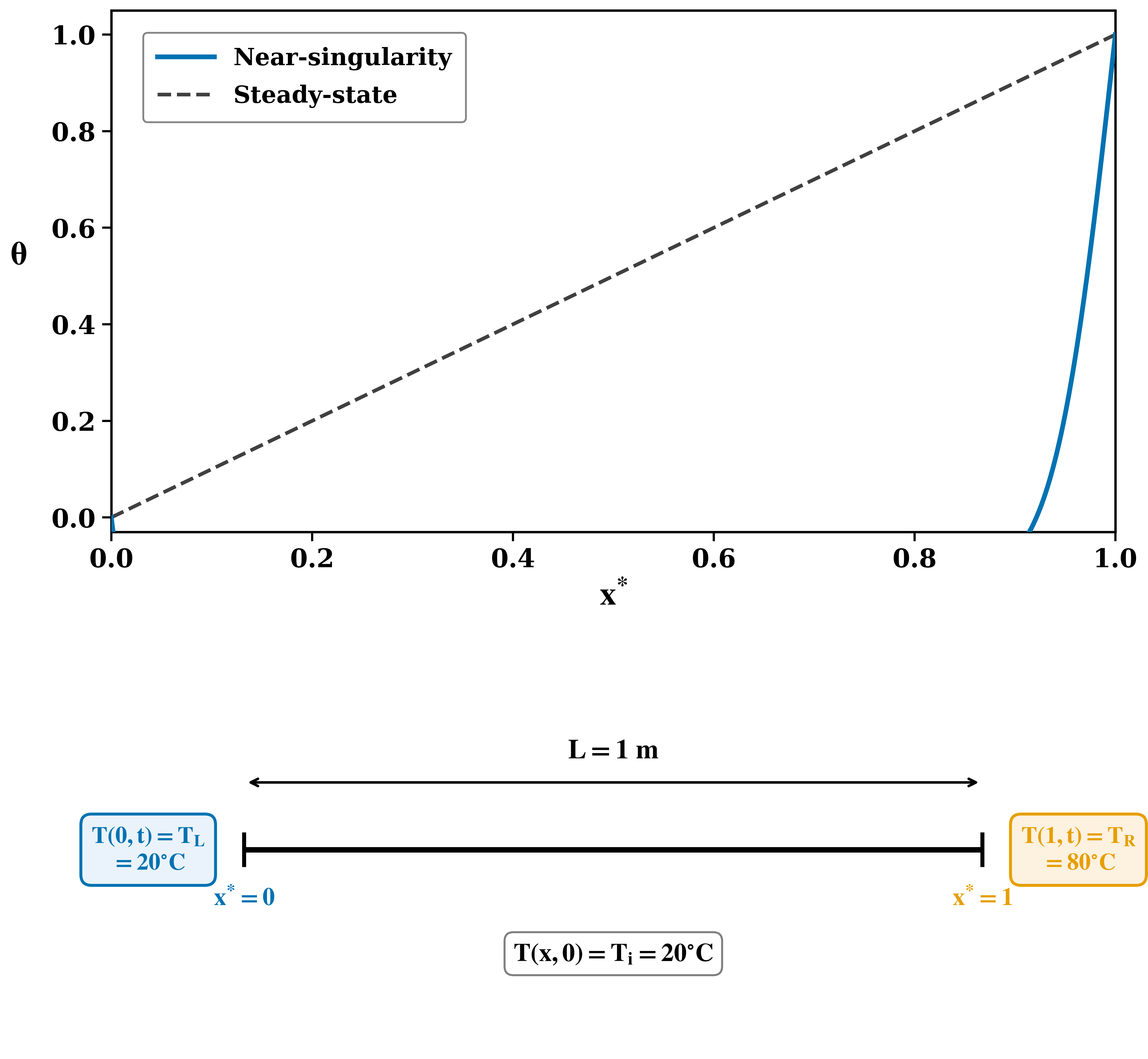}
\caption{Physical configuration of the analytical benchmark: a step
change in boundary temperature whose closed-form solution supplies
exact, unlimited ground truth for certifying extrapolated predictions
at every Fourier number.}
\label{fig:schematic_1d}
\end{figure}

To provide a benchmark for developing and validating machine-learning 
extrapolation strategies, a one-dimensional transient diffusion problem
is adopted as a controlled testbed --- selected not as the primary
engineering problem this paper sets out to solve, but because its
closed-form analytical solution makes every extrapolated prediction
checkable against an exact answer. Figure~\ref{fig:schematic_1d}
illustrates the configuration. A rod of length $L = 1\,\mathrm{m}$
with uniform thermal diffusivity
$\alpha = 1\times10^{-5}\,\mathrm{m}^2\,\mathrm{s}^{-1}$ is
initially at a uniform temperature of $T_i = 20\,^\circ\mathrm{C}$.
At $t = 0$, the right boundary is suddenly raised to
$T_R = 80\,^\circ\mathrm{C}$ while the left boundary is held at
$T_L = 20\,^\circ\mathrm{C}$. The problem admits an exact analytical
solution at every instant, which means that any prediction the model
makes can be verified precisely, regardless of how far that prediction
lies from the training window.

The governing equation is~\cite{Ozisik1993}
\begin{equation}
\frac{\partial T}{\partial t} = \alpha\,\frac{\partial^2 T}{\partial x^2},
\quad 0 < x < L,\;t > 0,
\label{eq:heat_dim}
\end{equation}
with boundary conditions $T(0,t)=T_L$, $T(L,t)=T_R$ for $t>0$, and
initial condition $T(x,0)=T_i$ for $0\le x<L$. At $t=0$ the field is
uniform and entirely discontinuous with the right-wall value, producing
a sharply nonlinear initial state that the benchmark must later require
a model to extrapolate toward.



Introducing the dimensionless coordinate $x^* = x/L$, dimensionless
temperature $\theta = (T - T_L)/(T_R - T_L)$, and Fourier number
$Fo = \alpha t/L^2$, Eq.~\eqref{eq:heat_dim} reduces to the
parameter-free form~\cite{Ozisik1993}
\begin{equation}
\frac{\partial\theta}{\partial Fo}
= \frac{\partial^2\theta}{\partial {x^*}^2},
\quad 0 < x^* < 1,\;Fo > 0,
\label{eq:heat_nodim}
\end{equation}
with $\theta(0,Fo)=0$, $\theta(1,Fo)=1$, and $\theta(x^*,0)=0$.
Because $Fo$ is the sole governing parameter, extrapolation
reliability certified against this dimensionless solution is
transferable to any physical system sharing the same $Fo$, regardless
of the specific diffusivity, length, or boundary temperatures.

\subsection{Analytical Solution}
\label{subsec:nondim}
The exact analytical solution is~\cite{Ozisik1993,Carslaw1959}:
\begin{equation}
\begin{aligned}
\theta(x^*,Fo)
&= x^*
 + \sum_{n=1}^{N_\mathrm{terms}}
   C_n \sin(n\pi x^*)\,e^{-n^2\pi^2 Fo}, \\[4pt]
C_n &= \frac{2}{n\pi}\bigl[(-1)^n - 1\bigr],
\end{aligned}
\label{eq:fourier_series}
\end{equation}
where $N_\mathrm{terms} = 1000$ and $Fo \geq 10^{-4}$. This closed
form is exact at every $Fo$ and can be evaluated at any spatial
location with no experimental or computational cost, supplying
unlimited ground-truth data for both training and validation. It is
this property---exact and freely available reference values
everywhere---that makes the diffusion problem suitable as a
validation benchmark rather than merely a convenient worked example.
\begin{figure}
\centering
\includegraphics[width=0.88\columnwidth]{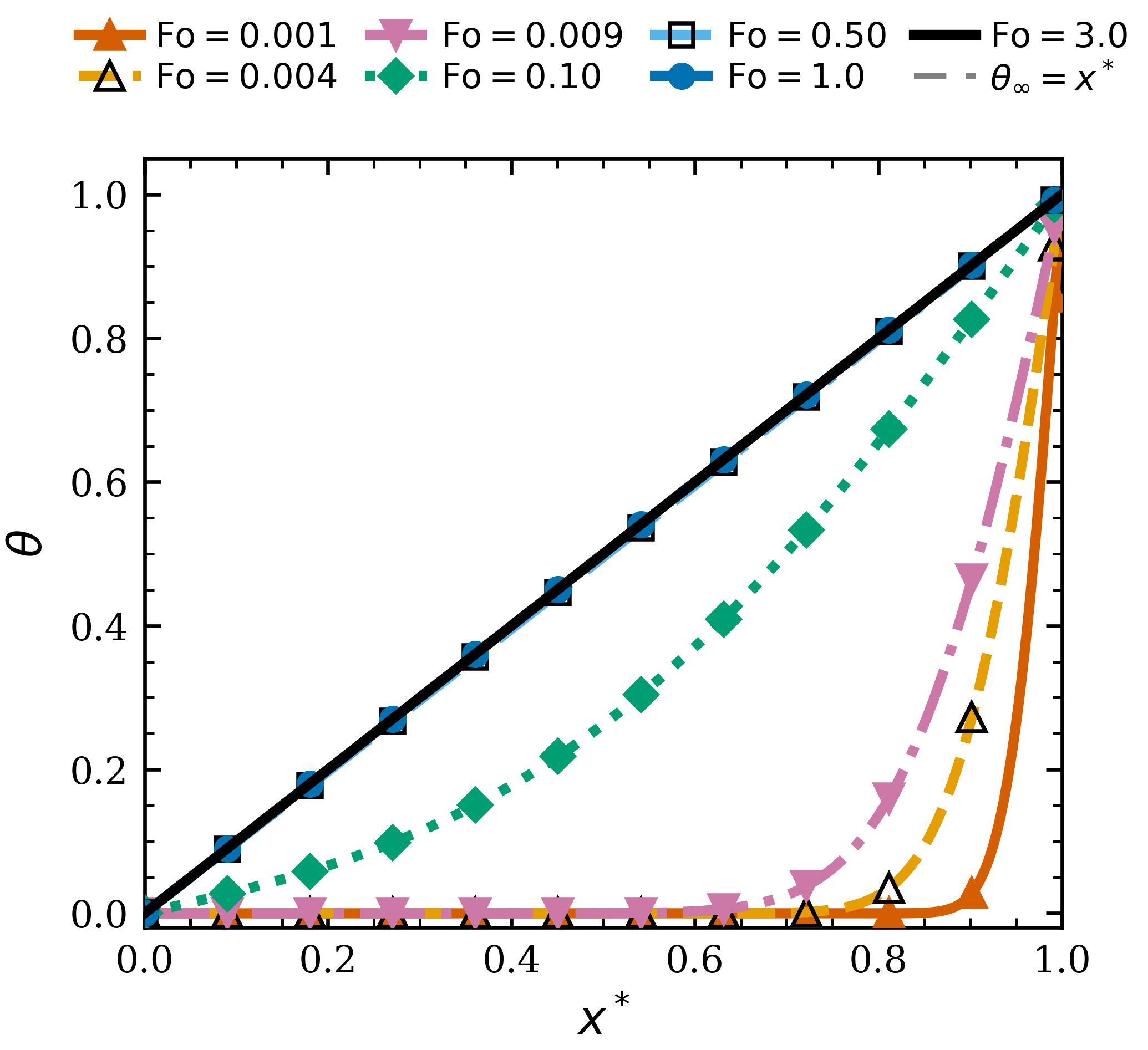}
\caption{Analytical solution of the transient diffusion benchmark at
representative Fourier numbers.}
\label{fig:analytical_profiles}
\end{figure}
Each term in Eq.~\eqref{eq:fourier_series} decays
according to~\cite{Ozisik1993}
\begin{equation}
A_n(Fo) = e^{-n^2\pi^2 Fo},
\label{eq:modal_decay}
\end{equation}
so higher-order modes ($n$ large) decay much faster than low-order
ones as $Fo$ increases; equivalently, at small $Fo$ a correspondingly
larger number of modes remain significant. The early-time asymptotic
solution of the diffusion equation shows that the temporal rate of
change scales as~\cite{Ozisik1993}
\begin{equation}
\left|\frac{\partial\theta}{\partial Fo}\right| \propto Fo^{-3/2},
\label{eq:gradient_singularity}
\end{equation}
so the solution changes increasingly rapidly as $Fo \to 0$.

Figure~\ref{fig:analytical_profiles} shows the solution evolution
across this range. Because the modal amplitudes in
Eq.~\eqref{eq:fourier_series} decay as
$e^{-n^2\pi^2 Fo}$ (Eq.~\eqref{eq:modal_decay}), higher-order Fourier
modes remain significant at small $Fo$, producing steep profiles with
rapid spatial and temporal variation. As $Fo\to0$, the rate of temporal
variation becomes singular according to
Eq.~\eqref{eq:gradient_singularity}. With increasing $Fo$, the
higher-order modes progressively decay, and the solution relaxes
smoothly toward the linear steady-state profile
$\theta_\infty=x^*$. This evolution provides distinct extrapolation
regimes, ranging from the rapidly varying early-time solution to the
smooth, low-mode-content regime at larger $Fo$
\cite{Ozisik1993,Carslaw1959}. The dashed line, $\theta = x^*$, represents the steady state.

\subsection{Extrapolation Validation Strategy}
\label{subsec:analytical}

Machine-learning models deployed in engineering practice are typically
trained over a limited operating range, while data outside that range
are rarely available for validation. This makes it difficult to
establish whether a model can be trusted beyond the conditions on which
it was trained. The one-dimensional transient diffusion problem is
adopted here to overcome this limitation. Its closed-form solution
(Eq.~\eqref{eq:fourier_series}) provides exact reference values at
every Fourier number and spatial location, allowing extrapolated
predictions to be assessed against known ground truth rather than
through visual comparison or against another surrogate model.

The model is deliberately trained only over the intermediate range
$Fo \in [0.001,,0.009]$ and evaluated outside this window in both
directions. Forward extrapolation moves toward a slowly varying,
low-mode-content regime, whereas backward extrapolation toward
$Fo\rightarrow0$ requires recovery of increasingly fine spatial and
temporal structure. The latter therefore provides a more demanding
test of extrapolation capability.

The exact analytical solution enables a sequential
train--predict--validate--extend strategy in which extrapolation is
verified at every step before advancing further. Specifically, the
procedure consists of five stages:

\begin{enumerate}
\item \textbf{Training:} The model is trained exclusively on data
within the prescribed Fourier-number window,
$Fo \in [0.001,0.009]$.

\item \textbf{Prediction:} The trained model predicts the solution
at the next Fourier number outside the training range.

\item \textbf{Validation:} The predicted solution is compared
against the exact analytical solution, providing a direct
pointwise measure of extrapolation error.

\item \textbf{History update:} Once the prediction satisfies the
prescribed validation criterion, it is incorporated into the
model's working history as the next available point.

\item \textbf{Extension:} The prediction horizon is advanced by one, further step, and the predict--validate--update cycle is repeated
until the desired extrapolation range is reached.

\end{enumerate}

This sequential procedure emulates extrapolative deployment while
retaining exact ground-truth verification at every stage. It therefore
provides a controlled and fully quantifiable framework for assessing
how reliably the Physics-guided BiLSTM and PINN models extrapolate as
they move progressively farther from their training domain.

\subsection{Analytical Dataset \& Training}
\label{subsec:datagen}

The analytical solution~\eqref{eq:fourier_series} is evaluated over the
entire heating process, from the initial temperature jump to the
near-steady-state condition, covering
$Fo \in [0.0005,,4.0]$. The spatial domain is discretized using
$N_x=401$ points along the rod, corresponding to
$\Delta x^*=0.0025$. This complete dataset provides exact reference
solutions for evaluating model predictions both within and outside the
training range.

The Fourier-number sampling is non-uniform to account for the
different rates of solution evolution. At small $Fo$, the solution
exhibits rapid temporal variation and steep gradients near the rod
ends, requiring finer sampling. As the solution progressively
smooths toward steady state, the sampling interval is increased. The
resulting $\Delta Fo$ values are $0.0005$, $0.01$, $0.02$, and $0.1$
over successive regions of the Fourier-number domain, concentrating
data where the solution changes most rapidly.

For model development, only a narrow subset of this dataset,
$Fo\in[0.001,,0.009]$, is used. This window encompasses a strongly
transient regime in which several Fourier modes remain active while
avoiding the most singular early-time behaviour. Within this window,
the solution is uniformly resampled with
$\Delta Fo=10^{-4}$, yielding $N_{Fo}=81$ Fourier-number values and a
total of $81\times401=32{,}481$ spatio-temporal data points. These are
split into training and validation sets using an $85\%/15\%$ ratio
with fixed random seeds.

Model performance is assessed at two levels. First, interpolation is
tested at $Fo=0.002$ and $Fo=0.007$, which lie within the training
window and provide a basic consistency check. The primary assessment
then considers extrapolation beyond the training domain. The first
extrapolation points are $Fo=0.0009$ and $Fo=0.0091$, immediately
outside the lower and upper training boundaries, respectively. The
prediction horizon is subsequently extended step by step in both
directions. At each step, the prediction is compared against the exact
analytical solution and, once validated, incorporated into the
working history before advancing to the next Fourier number. This
sequential procedure enables extrapolation accuracy to be evaluated
as the model moves progressively farther from its training domain.

$R^2$ scores were used as the scoring metric. Model performance was evaluated on the held-out test set using the mean absolute error (MAE), root mean square error (RMSE), and $R^2$ score, which are defined as:
\begin{equation}
\mathrm{MAE}  = \frac{1}{n} \sum_{i=1}^{n} \left| y_i - \hat{y}_i \right|, \hspace{1mm}
R^2  = 1 - \frac{\sum_{i=1}^{n} \left( y_i - \hat{y}_i \right)^2}{\sum_{i=1}^{n} \left( y_i - \bar{y} \right)^2},
\end{equation}

\noindent
where  $y_i$ is the true value,  $\hat{y}_i$ is the predicted value,  $\bar{y}$  is the mean of the true values, and  \textit{n}  is the total number of samples.

\subsection{Temporal Marching Procedure}
\label{subsec:marching_strategy}

A model is not required to predict a distant Fourier number in a
single step. Instead, both models reach their target through a
sequence of incremental predictions, where each new prediction is
generated using the model's updated working history. In this
benchmark, each intermediate prediction can additionally be checked
against the exact analytical solution before proceeding to the next
step.

Starting from the last known point at the boundary of the training
range, the model predicts the temperature field at the next Fourier
number. The predicted field is then incorporated into the working
history as though it were a known observation, and the updated history
is used to generate the subsequent prediction. This process is
repeated until the target Fourier number is reached. For example,
forward extrapolation proceeds as
$Fo=0.0091\rightarrow0.0095\rightarrow0.010\rightarrow0.011
\rightarrow0.012$, while backward extrapolation follows
$Fo=0.0009\rightarrow0.00085\rightarrow0.00080\rightarrow0.00075$.
The same sequential procedure is applied to the additional early-time
targets discussed in Section~\ref{sec:results}.

This setup reflects how an extrapolative model would operate in
practice: beyond the training domain, each prediction depends on the
model's previously generated outputs rather than on newly supplied
ground-truth data. The analytical solution is used here only to
evaluate each step and quantify error accumulation. Thus, the
benchmark provides a controlled means of assessing how prediction
reliability deteriorates as the model progressively moves beyond its
training domain.

\section{Demonstration I: Physics-Guided Sequential Extrapolation Framework Using BiLSTM}
\label{sec:bilstm}

This section uses a sequential-learning architecture as the first of
two demonstrations of the validation framework set out in
Section~\ref{sec:governing}. The question is whether a sequential
architecture can be modified using physical insight to extrapolate
reliably beyond the training domain, not whether a BiLSTM can solve
heat conduction inside its training window; Section~\ref{sec:results}
shows that even the standard model manages that comfortably.
Section~\ref{sec:governing} established three facts relevant here:
time behaves like $\sqrt{Fo}$, not like $Fo$; the steepest gradients
and richest mix of Fourier modes sit near the boundaries; and
stepping backward to smaller $Fo$ works against the natural smoothing
effect of diffusion rather than with it. A Bi-LSTM accounts
for none of this: it reads $Fo$ as an ordinary number, treats every
spatial point alike, and feeds its own output back in at each step
with no check on how errors might grow. The Physics-guided BiLSTM
below addresses these three extrapolation barriers directly, one
targeted correction for each, so that its extrapolated predictions
can pass the validation test of Section~\ref{subsec:analytical}.

These corrections are derived from the benchmark physics itself,
not introduced empirically. The rapid early-time variation near
$Fo \to 0$ established in Section~\ref{sec:governing} motivates the
Root-Fourier transformation, which realigns the network's notion of
time with the true diffusion timescale rather than the linear $Fo$
the standard model reads. The strong gradients concentrated at the
domain boundaries motivate the boundary-aware, boundary-weighted
loss, which directs the network's attention to the region where
errors matter most. The asymmetry between backward and forward
extrapolation established in Section~\ref{sec:governing} motivates
the stability-enhancing relaxation marching, which bounds error
growth specifically in the harder backward direction. Each
modification therefore exists to correct a known consequence of the
benchmark physics, not as an arbitrary architectural choice.

\subsection{Bi-LSTM}
\label{subsec:lstm_math}

At each time step $k$, the LSTM cell updates its hidden state
$\mathbf{h}_k$ and cell state $\mathbf{c}_k$ through four gating
operations~\cite{Hochreiter1997}:
\begin{equation}
\begin{aligned}
\mathbf{f}_k &=
\sigma\!\left(W_f[\mathbf{x}_k;\,\mathbf{h}_{k-1}] + b_f\right),
  \label{eq:forget_gate} \\
\mathbf{i}_k &=
\sigma\!\left(W_i[\mathbf{x}_k;\,\mathbf{h}_{k-1}] + b_i\right),
   \\
\tilde{\mathbf{c}}_k &=
\tanh\!\left(W_c[\mathbf{x}_k;\,\mathbf{h}_{k-1}] + b_c\right),
 \\
\mathbf{o}_k &=
  \sigma\!\left(W_o[\mathbf{x}_k;\,\mathbf{h}_{k-1}] + b_o\right).
\end{aligned}
\end{equation}
In the BiLSTM, two independent LSTM chains process the input sequence
in opposite temporal directions, and their hidden states are
concatenated at each step~\cite{Schuster1997,Graves2013}.

\subsection{Limitations of the Bi-LSTM in the Early-Time Regime}
\label{subsec:lstm_limitations}

The gates above take $Fo$ as a plain input, but the benchmark
solution evolves on a $\sqrt{Fo}$ timescale: near $Fo \to 0$, equal
steps in $Fo$ correspond to very unequal amounts of physical change,
so a model reading $Fo$ on a straight line already carries a distorted
sense of time exactly where it can least afford to.

A second problem follows from where the hardest part of the solution
sits: the boundary near $x^* \to 1$ has the steepest gradients and
the most rapidly varying profile when $Fo$ is small. A Bi-LSTM
has no built-in reason to pay extra attention there, since its loss
treats every spatial point alike.

The third problem appears once prediction moves outside the training
window. Reaching a target below the training range means marching
backward step by step (Section~\ref{subsec:marching_strategy}), and,
as Section~\ref{sec:governing} showed, this direction works against
diffusion's natural smoothing rather than with it. A Bi-LSTM
has no way to stop a small error at one step from growing into the
next, a real concern given that different spatial modes are
amplified unequally during backward marching.
These three barriers, the wrong notion of time, no extra care at the
boundary, and no control over error growth during backward marching,
follow directly from the physics in Section~\ref{sec:governing}.


\subsection{Physics-guided BiLSTM Architecture}
\label{subsec:architecture}

\begin{figure}
\centering
\includegraphics[width=0.75\columnwidth]{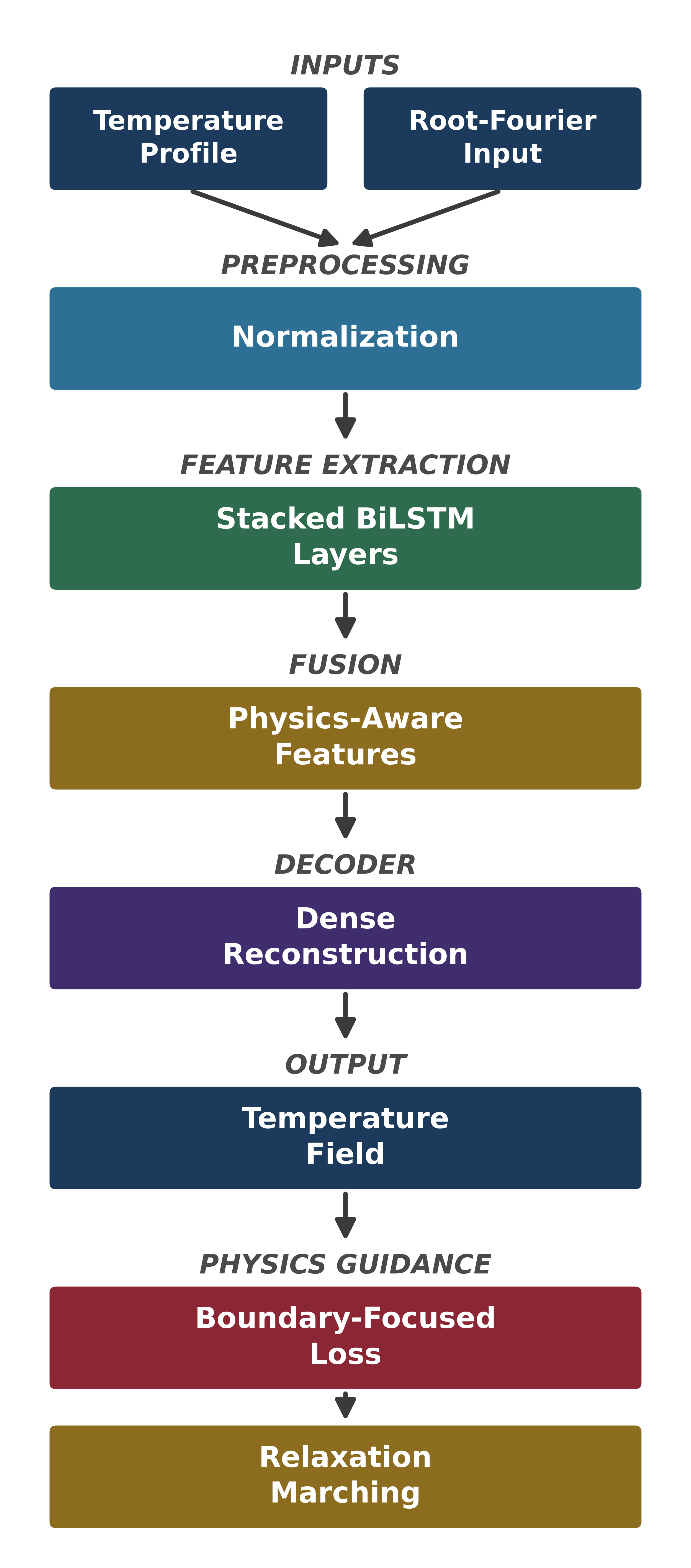}
\caption{Physics-guided BiLSTM framework, the first of two independent
demonstrations of the extrapolation-validation methodology: a
Root-Fourier input ($s=\sqrt{Fo}$) realigns the network's time
coordinate with the diffusion timescale, a boundary-weighted loss
targets the highest-curvature region, and relaxation marching bounds
error growth during sequential extrapolation.}
\label{fig:bilstm_arch}
\end{figure}

The changes below are targeted corrections for the three barriers
above, not a general accuracy improvement: the goal is a better match
between architecture and physics, not added complexity. Three stacked
bidirectional LSTM layers process input sequences $(x^*, s, \theta)$
over a sliding window of width $K = 8$, with a four-layer fully
connected decoder reconstructing the full spatial temperature profile
(Figure~\ref{fig:bilstm_arch}). All models are implemented in PyTorch
and trained on an NVIDIA Tesla T4 GPU with the Adam optimiser, initial
learning rate $10^{-3}$, batch size $16$, early stopping with patience
$100$, and a \texttt{ReduceLROnPlateau} scheduler (reduction factor
$0.5$, patience $30$). The total parameter count is $422{,}801$.

\subsubsection{Root-Fourier Transformation}
\label{subsec:rootfo}

The first barrier is the mismatch between the model's plain $Fo$
input and the real physical clock: thermal penetration depth grows
with $\sqrt{Fo}$, not $Fo$ itself~\cite{Incropera2007,Carslaw1959}, so
equal steps in $Fo$ stop meaning equal physical change exactly as
$Fo \to 0$. The correction aligns the learning coordinate with the
true diffusion timescale. The Root-Fourier coordinate
$
s \equiv \sqrt{Fo}
$
straightens out the penetration-depth relationship and softens the
temporal singularity from $Fo^{-3/2}$ to the milder $Fo^{-1/2}$. In
$s$-coordinates, the training range is $s \in [0.001,\,0.009]$ (in
$\sqrt{Fo}$), and the marching step from $Fo = 0.001$ to
$Fo = 0.0009$ corresponds to $\Delta s \approx 0.0016$, only $5\%$ of
this range. The model's extrapolation input therefore changes
smoothly in step with the underlying physics, rather than squeezing a
large physical change into a tiny numerical step.

\subsubsection{Boundary-Weighted Loss}
\label{subsec:bwloss}

The second barrier is that nothing tells the model the boundary
matters more. The spatial curvature of the $n$th Fourier eigenfunction
$\sin(n\pi x^*)$ is largest near the domain boundaries, and these
high-curvature near-boundary modes are amplified most strongly during
sequential marching. A loss weighing every point equally
under-values exactly the region Section~\ref{sec:governing} flagged
as hardest. The boundary-weighted loss~\cite{mcclenny2023,wight2021}
corrects this:
\begin{equation}
\mathcal{L}_{BW} =
\frac{1}{N_x N_{Fo}}
\sum_{j=1}^{N_x}\sum_{k=1}^{N_{Fo}}
w(x^*_j)\,
\bigl[\hat{\theta}(x^*_j,Fo_k) -
      \theta(x^*_j,Fo_k)\bigr]^2,
\label{eq:bw_loss}
\end{equation}
with spatially adaptive weight
\begin{equation}
\begin{aligned}
w(x^*) &= 1 + A
\left[
e^{-{x^*}^2/(2\sigma_w^2)}
+ e^{-(1-x^*)^2/(2\sigma_w^2)}
\right], \\
&\quad A = 5,\quad \sigma_w = 0.1.
\end{aligned}
\label{eq:weight_function}
\end{equation}
This gives $w(0) = w(1) = 6.0$ at the boundaries and
$w(0.5) \approx 1.0$ at the midpoint, penalising boundary errors six
times more heavily than interior errors and steering the network
toward learning the hardest part of the physics, precisely where
extrapolation fidelity would otherwise be lost first.

\subsubsection{Relaxation Marching}
\label{subsec:relaxation}

The third barrier appears once the model steps outside its training
window. In sequential marching (Section~\ref{subsec:marching_strategy}),
each new prediction is fed back in to make the next, so any error
made at one step becomes part of the input to the next. This is
riskier backward, where diffusion no longer damps such errors and can
amplify them: each Fourier mode decays by a factor $e^{-n^2\pi^2 Fo}$
moving forward in $Fo$, so moving backward inverts this factor and
grows the mode instead, with higher-order modes amplified most. Relaxation marching controls this
error accumulation by blending each new prediction partway with the
previous state rather than accepting it outright, applying an
under-relaxation update at each sub-step~\cite{Patankar1980}:
\begin{equation}
\theta^{(k+1)}(x^*) = \theta^{(k)}(x^*) +
\omega\!\left[\hat{\theta}^{(k+1)}(x^*) - \theta^{(k)}(x^*)\right],
\label{eq:relaxation}
\end{equation}
where $\omega \in (0,1]$ is the under-relaxation factor, and the
relaxed convergence factor is
\begin{equation}
\rho = 1 + \omega(\lambda - 1).
\label{eq:rho_formula}
\end{equation}
With $\lambda = 1.001$ and $\omega = 0.3$:
\begin{equation}
\rho_\mathrm{early} = 0.7 + 0.3 \times 1.001 = 1.0003.
\label{eq:rho_back}
\end{equation}
Over six sub-steps, cumulative error growth is
$\rho_\mathrm{early}^6 \approx 1.0018$, a small, known number rather
than an open-ended one, keeping errors small and predictable even
when marching backward, where diffusion offers no such guarantee on
its own. Convergence parameters for both directions are summarised in
Table~\ref{tab:relaxation}.
\begin{table}
\centering
\footnotesize
\setlength{\tabcolsep}{4pt}
\renewcommand{\arraystretch}{1.05}
\caption{Relaxation marching convergence parameters for backward and
forward temporal directions.}
\label{tab:relaxation}
\begin{threeparttable}
\begin{tabular}{@{}lccccc@{}}
\toprule
Direction & $\omega$ & $\lambda$ & $\rho$ & $N$ & $\rho^{N}$ \\
\midrule
Backward ($Fo$ decreasing) & 0.30 & 1.001 & 1.0003 & 6 & 1.0018 \\
Forward\, ($Fo$ increasing) & 0.70 & 0.999 & 0.9993 & 4 & 0.9972 \\
\bottomrule
\end{tabular}
\begin{tablenotes}\footnotesize
\item $\omega$: under-relaxation factor; $\lambda$: raw modal error growth;
      $\rho=1+\omega(\lambda-1)$: relaxed factor; $N$: sub-steps.
\end{tablenotes}
\end{threeparttable}
\end{table}

\subsection{Role of the Physics-guided BiLSTM Framework in Extrapolation Validation}
\label{subsec:bilstm_role}

The Physics-guided BiLSTM here is one implementation of the
validation framework set out in Section~\ref{subsec:analytical}, not the
contribution that framework exists to support. The framework itself,
restricting the training domain, marching sequentially beyond it, and
checking every step against exact ground truth, stays unchanged
regardless of which architecture is plugged into it; what BiLSTM
contributes is a concrete demonstration that a sequential, data-driven
architecture can be corrected, using only physical insight into the
governing equation, until its extrapolated predictions reliably pass
that validation test. The next section repeats this same assessment
with a second architecture built on entirely different principles.

\section{Demonstration II: Modified PINN Extrapolation Framework }
\label{sec:pinn}

The BiLSTM tested one route to extrapolation reliability: marching
through labelled data step by step, with validation against the
analytical solution at every step. The purpose of introducing a PINN
is not to provide an alternative heat-conduction solver, but to test
whether a fundamentally different learning paradigm can also achieve
reliable extrapolation under the same validation framework. Instead
of relying only on data, it builds the governing equation directly
into the loss function, so the model can be pushed toward a correct
answer even at points with no labelled data~\cite{Raissi2019},
including points beyond $Fo = 0.009$. This is appealing here, since
it offers physically sensible answers exactly where a purely
data-driven model would have nothing nearby to march from. Before
this advantage can be used reliably, two extrapolation barriers that
follow directly from the physics in Section~\ref{sec:governing} need
to be addressed first.

The same benchmark characteristics that motivated the BiLSTM
corrections create a distinct difficulty for a PINN. Early-time
behaviour near $Fo \to 0$ is highly nonlinear, and the solution
changes increasingly rapidly as $Fo$ decreases, which is precisely
what makes the residual term in the PINN's loss function increasingly
difficult to optimise as the prediction target approaches the
near-singular regime: gradients that ought to guide training instead
grow without bound. Hard-enforced boundary conditions face a related
problem, since approximate enforcement leaves room for boundary error
to leak into the interior exactly where the solution is least
forgiving. The physics-guided coordinate transformation and hard
boundary constraints introduced below exist specifically to remove
these two difficulties at their source, rather than to compensate for
them empirically during training.

\subsection{PINN and Limitations}
\label{subsec:pinn_standard}

A PINN predicts $\hat{\theta}(x^*, Fo)$ by minimising a composite
loss~\cite{Raissi2019,lagaris1998}, built around the PDE residual
\begin{equation}
\mathcal{R}(x^*, Fo) =
\frac{\partial\hat{\theta}}{\partial Fo}
- \frac{\partial^2\hat{\theta}}{\partial {x^*}^2}.
\label{eq:pde_residual}
\end{equation}
This residual lets the network train without labelled answers
everywhere: wherever it can be computed, the network is nudged toward
a physically valid solution, whether or not the true value of
$\theta$ is known there.

Two features of the physics make this harder near $Fo \to 0$. As
shown in Section~\ref{sec:governing}, the gradient singularity means
$\partial\theta/\partial Fo$ scales as $Fo^{-3/2}$, so the residual
becomes far harder to compute and minimise reliably in this limit,
and one fixed learning rate cannot work well both there and in the
rest of the domain~\cite{Wang2021ntk,Raissi2019}. The second issue is
that a PINN only encourages $\theta(0,Fo)=0$ and
$\theta(1,Fo)=1$ through a soft penalty, so some boundary error always
remains; since the boundary carries the largest curvature and
gradients, this error spreads into the interior through the diffusion
operator itself~\cite{lagaris1998,mcclenny2023}. Both problems point
to the same conclusion: the time coordinate the PDE is written in,
and how the boundary conditions are enforced, both need to change
before a PINN can be trusted in the near-singular early-time regime.

\subsection{Network Architecture and Training Configuration}
\label{subsec:pinn_arch}

\begin{figure*}
\centering
\includegraphics[width=0.99\textwidth]{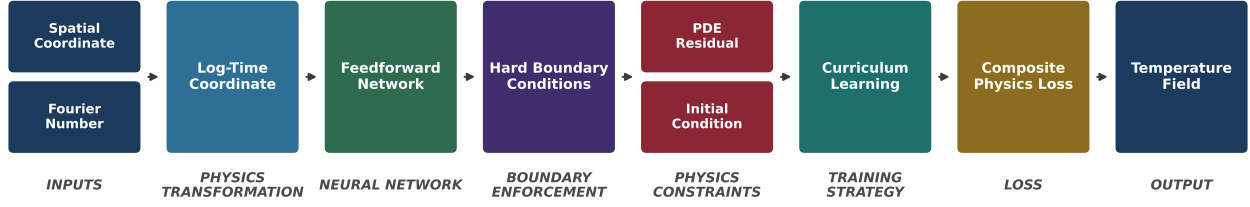}
\caption{Modified PINN framework, the second and independent
demonstration of the extrapolation-validation methodology: a
log-time coordinate removes the $Fo^{-3/2}$ residual singularity,
hard boundary-condition enforcement eliminates near-wall error
leakage, and curriculum Fourier-number expansion stages training
toward the most demanding extrapolation targets.}
\label{fig:pinn_arch}
\end{figure*}

The PINN comprises six fully connected hidden layers with $60$
neurons each and $\tanh$ activations ($18{,}541$ total parameters),
illustrated in Figure~\ref{fig:pinn_arch}. Training runs over
$20{,}000$ epochs with a piecewise-constant learning rate schedule;
reductions at epochs $12{,}000$ and $16{,}000$ allow fine-tuning in
the late stages.

\subsection{Modified PINN Corrections}

The two barriers above need two separate corrections. The gradient
singularity is addressed first, by changing the time coordinate so
the stiffness disappears at the source. The boundary-error barrier is
addressed second, by changing the architecture so the boundary
conditions hold exactly, freeing the optimiser to spend all its
effort on learning the PDE itself.

\subsubsection{Log-Time Transformation}
\label{subsec:logt}

The trouble near $Fo \to 0$ is that the residual gradient the
optimiser must minimise grows without limit as $\sim Fo^{-3/2}$, so no single fixed learning
rate can work well both in the bulk of the domain and near the
initial condition. The correction chooses a time coordinate where
this blow-up is tamed. The log-time coordinate
\begin{equation}
\tau = \log(Fo + \varepsilon),
\qquad \varepsilon = 10^{-5},
\label{eq:logt}
\end{equation}
does exactly that. Since $d\tau/dFo \approx 1/Fo$, the chain rule
gives
\begin{equation}
\frac{\partial\theta}{\partial Fo} =
\frac{1}{Fo+\varepsilon}\,
\frac{\partial\theta}{\partial\tau},
\label{eq:chain_rule}
\end{equation}
and substituting into Eq.~\eqref{eq:heat_nodim} yields
\begin{equation}
\frac{\partial\theta}{\partial\tau} =
(Fo+\varepsilon)\,
\frac{\partial^2\theta}{\partial{x^*}^2}.
\label{eq:transformed_pde}
\end{equation}
The prefactor $(Fo+\varepsilon) \rightarrow 0$ as $Fo \rightarrow 0$
analytically cancels the singularity: the residual gradient now
scales as $(Fo+\varepsilon)\,Fo^{-3/2} \approx Fo^{-1/2}$, one full
power of $Fo$ milder than the original $Fo^{-3/2}$. As $Fo$ drops from $0.01$ to $0.0001$, the untransformed gradient
grows by six orders of magnitude, while the transformed one grows by
only two. This leaves the loss
landscape near $Fo \to 0$ far better
conditioned~\cite{Wang2021ntk}, letting the network converge close to
the initial condition and extrapolate into it, where the
untransformed version struggles most.

\subsubsection{Hard Boundary Condition Enforcement}
\label{subsec:hardbc}

The second barrier is at the boundaries themselves. Because the
boundary layer carries the steepest gradients and richest mix of
modes (Section~\ref{sec:governing}), any error left there by a soft
penalty leaks into the interior through diffusion, contaminating the
extrapolated predictions this paper sets out to validate. The
correction makes it impossible for the boundary conditions to be
wrong in the first place, through a simple architectural split,
\begin{equation}
\theta(x^*,\tau) =
g(x^*) + h(x^*)\,\mathcal{N}(x^*,\tau),
\label{eq:hard_bc_decomp}
\end{equation}
where $g(x^*) = x^*$ and $h(x^*) = x^*(1-x^*)$, giving
\begin{equation}
T(x,Fo) =
T_L + x^*(T_R - T_L) + x^*(1-x^*)\,\mathcal{N}(x^*,\tau).
\label{eq:hard_bc}
\end{equation}
This enforces $T(0) = T_L$ and $T(1) = T_R$ exactly for any network
output $\mathcal{N}$, without relying on the optimiser to drive a
penalty term toward zero. In a standard soft-penalty
setup~\cite{Raissi2019,lagaris1998}, training effort always goes
toward shrinking the boundary error, and whatever remains still leaks
inward through the PDE coupling~\cite{Wang2021ntk,mcclenny2023}. The
hard-BC ansatz removes that cost entirely, freeing the optimiser to
focus on the PDE and initial-condition terms, which matters most
where the $Fo^{-3/2}$ singularity is worst.

\subsubsection{Composite Loss Function}
\label{subsec:compositeloss}

\begin{equation}
\mathcal{L} =
\lambda_\mathrm{data}\,\mathcal{L}_\mathrm{data}
+ \lambda_\mathrm{PDE}\,\mathcal{L}_\mathrm{PDE}
+ \lambda_\mathrm{IC}\,\mathcal{L}_\mathrm{IC},
\label{eq:composite_loss}
\end{equation}
with weights $\lambda_\mathrm{data} = 0.2$,
$\lambda_\mathrm{PDE} = 30$, and $\lambda_\mathrm{IC} = 50$. The
PDE residual loss is
\begin{equation}
\mathcal{L}_\mathrm{PDE} =
\frac{1}{N_r}\sum_{j=1}^{N_r}
\left[
\frac{\partial T_\theta}{\partial\tau_j}
- (Fo_j+\varepsilon)
  \frac{\partial^2 T_\theta}{\partial{x_j^*}^2}
\right]^{\!2}
\label{eq:pde_loss}
\end{equation}
evaluated over $N_r = 8{,}000$ collocation points per epoch,
including the region $Fo < 0.001$. The initial condition loss
$\mathcal{L}_\mathrm{IC}$ is evaluated over $N_\mathrm{IC} = 4{,}000$
points at $Fo \approx 0$.

\begin{table}
\centering
\footnotesize
\setlength{\tabcolsep}{4pt}
\renewcommand{\arraystretch}{1.05}
\caption{Modified PINN modifications and their physical rationale.}
\label{tab:pinn_mods}
\begin{tabularx}{\columnwidth}{@{}lXX@{}}
\toprule
\textbf{Modification} & \textbf{Limitation addressed} & \textbf{Expected benefit} \\
\midrule
Log-time transform
  & $Fo^{-3/2}$ residual singularity
  & Reduces singularity order to $Fo^{-1/2}$; improves gradient uniformity \\[3pt]
Hard BC
  & Soft-penalty boundary bias
  & Exact Dirichlet satisfaction; full optimisation budget for PDE/IC \\[3pt]
Curriculum learning
  & Training instability near $Fo\to0$
  & Progressive collocation expansion; avoids abrupt residual jumps \\
\bottomrule
\end{tabularx}
\end{table}

\begin{figure*}[!t]
\centering
\includegraphics[width=0.6\textwidth]{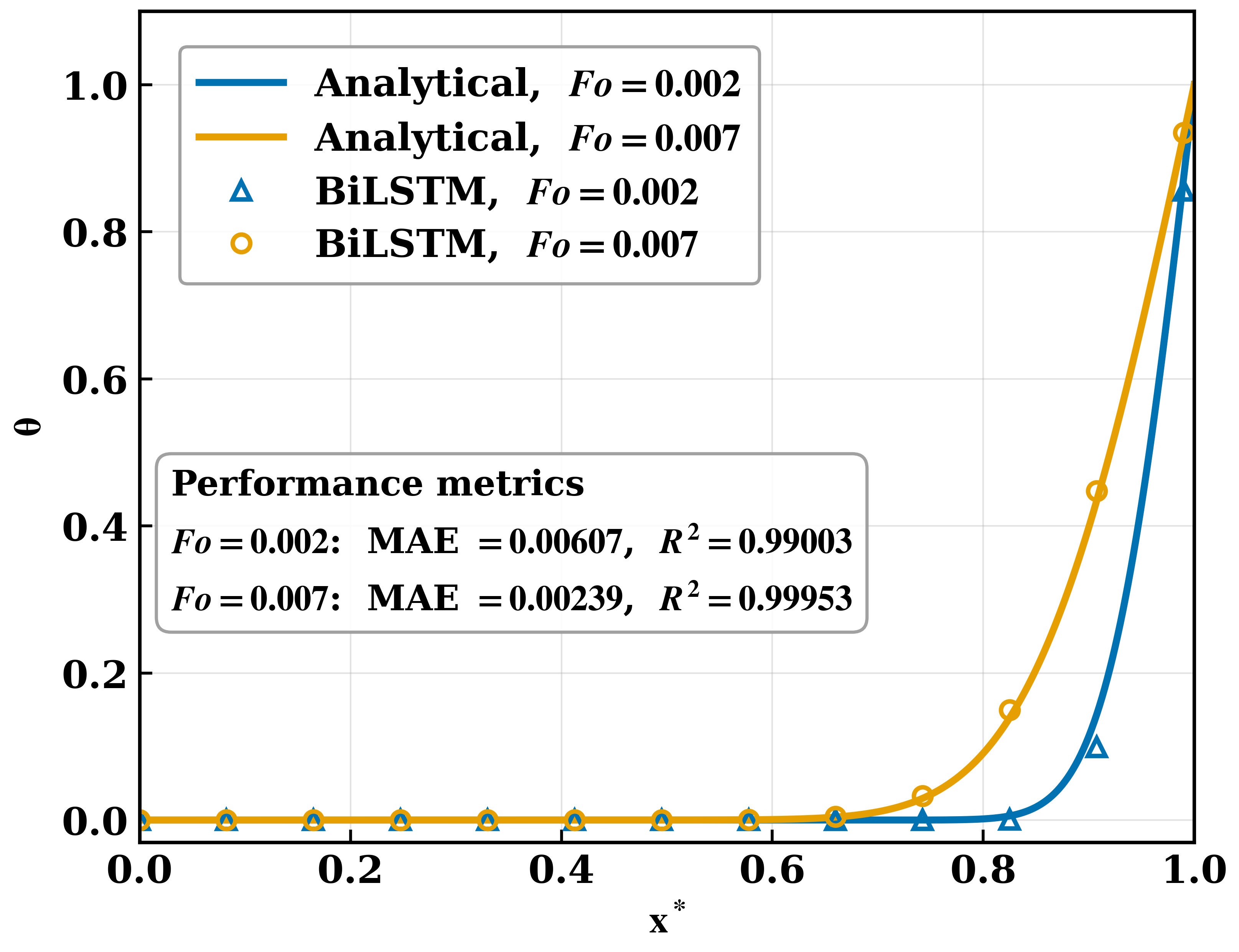}
\caption{Interpolation within the training range $Fo \in
[0.001,\,0.009]$ for the Physics-guided BiLSTM at (a)~$Fo=0.002$ and
(b)~$Fo=0.007$.}
\label{fig:bilstm_interpolation}
\end{figure*}

\begin{figure*}[!t]
\centering
\includegraphics[width=0.6\textwidth]{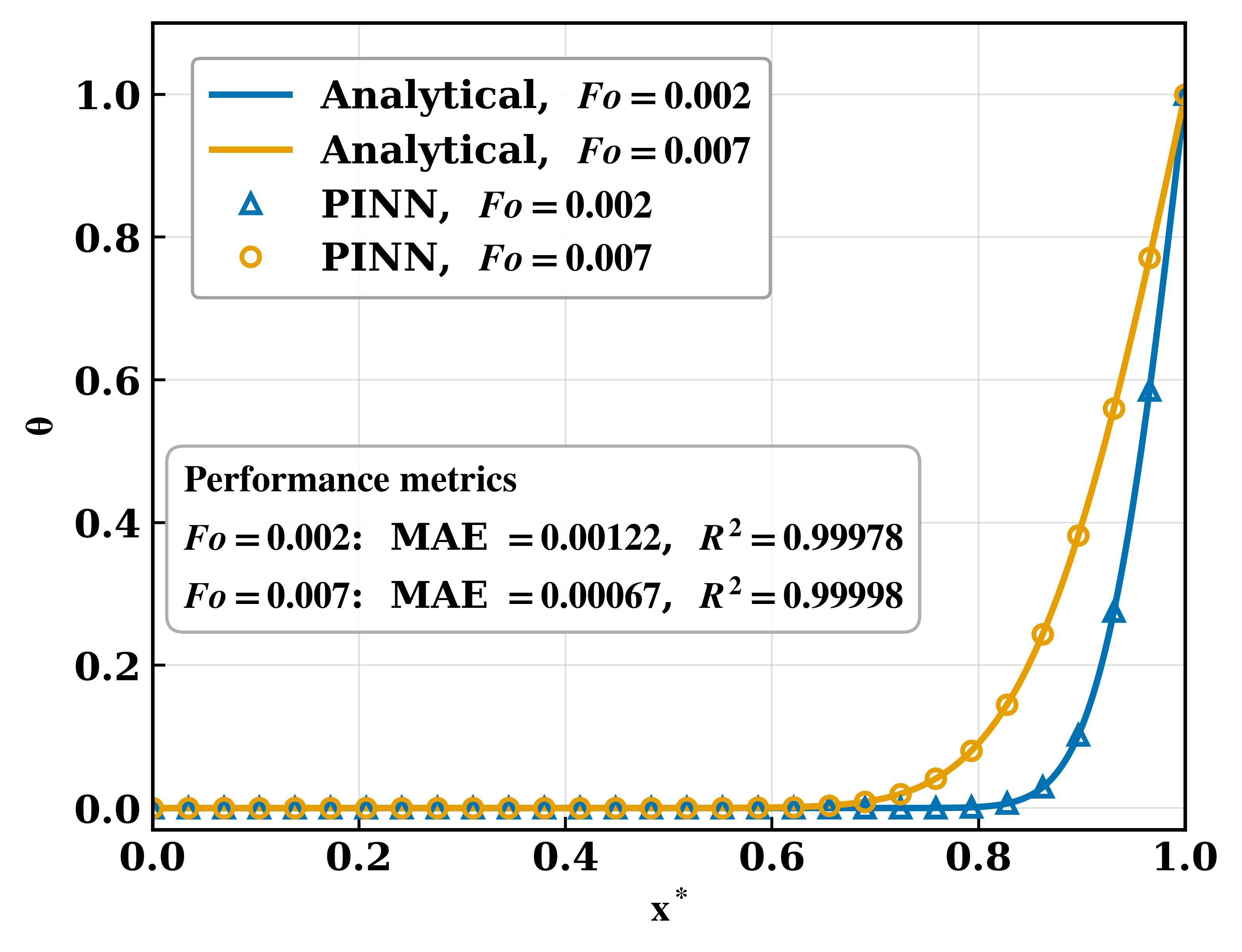}
\caption{Interpolation within the training range $Fo \in
[0.001,\,0.009]$ for the Modified PINN at (a)~$Fo=0.002$ and
(b)~$Fo=0.007$.}
\label{fig:pinn_interpolation}
\end{figure*}

\subsection{Role of the PINN Framework in Extrapolation Validation}
\label{subsec:pinn_role}

The Modified PINN serves as a second, independent validation
vehicle for the same framework demonstrated with the BiLSTM in
Section~\ref{sec:bilstm}: the methodology of Section~\ref{subsec:analytical}
did not change to accommodate a different learning paradigm. Where
the BiLSTM learns sequentially from labelled data, the PINN learns by
constraining itself to the governing equation where no labelled data
exist, yet the same train-then-validate procedure applies unchanged
to both. That the framework produces a reliably extrapolating model
under two architectures built on opposite principles is itself
evidence that the validation methodology is model-independent, which
is what makes it transferable to engineering problems where neither a
BiLSTM nor a PINN may be the right tool, but the same
train-extrapolate-validate logic still applies.

\section{Results and Discussions}
\label{sec:results}

This section presents the findings of the validation methodology described in Section~\ref{subsec:analytical}, with particular emphasis on the extrapolation reliability of the two frameworks. In-domain accuracy is established first, followed by an assessment of extrapolation performance across lower- and upper-regime Fourier numbers, with each prediction evaluated directly against the analytical ground truth. The frameworks are then compared in terms of their extrapolation behavior and engineering relevance. Throughout, the results are interpreted primarily as evidence of extrapolation robustness, rather than as findings specific to the heat-conduction testbed.

\begin{figure*}[!t]
\centering
\includegraphics[width=0.85\textwidth]{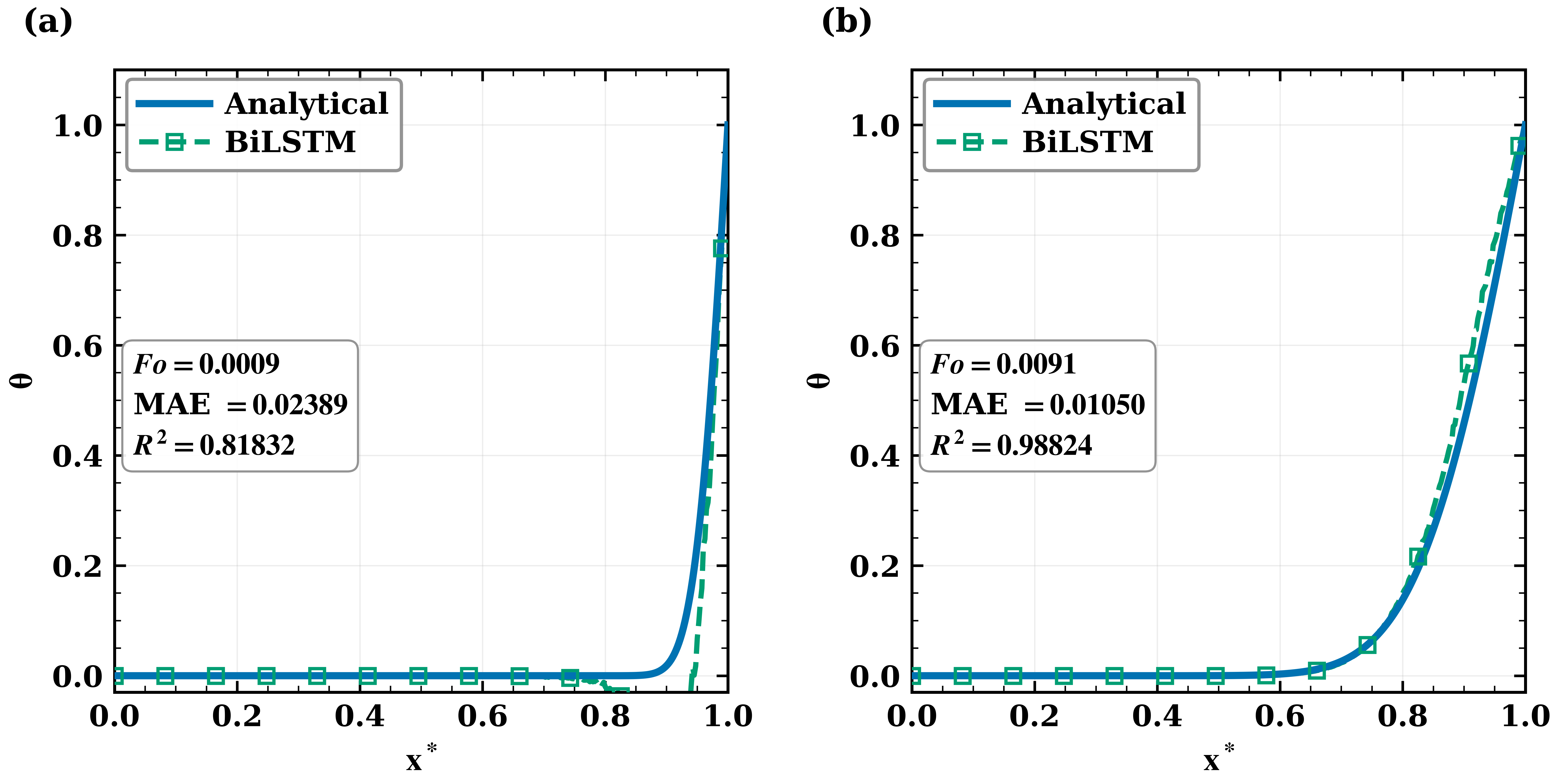}
\caption{Extrapolation from the training range $Fo \in
[0.001,\,0.009]$ using the Bi-LSTM to (a)~$Fo=0.0009$ and
(b)~$Fo=0.0091$, validated against the analytical solution.}
\label{fig:standard_lstm_extrapolation}
\end{figure*}
\subsection{Interpolation}
\label{subsec:interpolation}

Before evaluating extrapolation capability, both surrogate frameworks
must be confirmed to reproduce the governing diffusion physics within
the training window $Fo \in [0.001,\,0.009]$. Interpolation at two
representative Fourier numbers, $Fo = 0.002$ and $Fo = 0.007$, serves
this purpose, spanning the range from the spectrally richer lower
boundary, where more Fourier modes remain active, to the smoother
upper boundary, where high-frequency content has largely decayed. A
model that cannot reproduce the analytical solution inside its own
training window has little chance of extrapolating reliably outside
it, so this is a necessary checkpoint, not the main result, that
clears the way for the extrapolation tests that follow.

\subsubsection{BiLSTM Interpolation}
\label{subsubsec:bilstm_interp}

Figure~\ref{fig:bilstm_interpolation} reports the Physics-guided BiLSTM at both targets: $R^{2} = 0.99003$ with MAE $= 0.00607$ at
$Fo = 0.002$, tightening to $R^{2} = 0.99953$ with MAE $= 0.00239$ at
$Fo = 0.007$. The improvement from the lower to the upper target
matches the physical picture built up earlier: $Fo = 0.002$ still
carries a meaningful share of higher-order Fourier content near the
heated boundary, whereas $Fo = 0.007$ is dominated by the smoother
fundamental harmonic, an easier shape to reproduce. This confirms
baseline fidelity before any extrapolation claim is made, and previews
the same lower-versus-upper asymmetry that governs extrapolation
reliability later in this section.

\subsubsection{PINN Interpolation}
\label{subsubsec:pinn_interp}

Figure~\ref{fig:pinn_interpolation} shows the corresponding PINN
result: $R^{2} = 0.99978$ with MAE $= 0.00122$ at $Fo = 0.002$, rising
to $R^{2} = 0.99998$ with MAE $= 0.00067$ at $Fo = 0.007$. Both values
are tighter than the BiLSTM's fits at the same two Fourier numbers,
consistent with the PINN's boundary conditions being enforced exactly
rather than learned approximately, leaving less room for the kind of
near-wall error a purely data-driven recurrent model can accumulate.
Both frameworks thus establish the in-domain fidelity needed before
their extrapolation reliability can be tested outside the training
window.

\subsection{Extrapolation: BiLSTM}
\label{subsec:extrap_bilstm}

The next question this paper is built to answer: once each framework
predicts outside the training window altogether, do those predictions
hold up against the exact analytical ground truth, or fail silently
with no way for an engineer to know? Two opposite cases are tested: a
step backward to $Fo = 0.0009$, just below the lower training
boundary, and a step forward to $Fo = 0.0091$, just above the upper
boundary. The two directions are not expected to behave alike, since
backward extrapolation works against diffusion's natural smoothing,
while forward extrapolation is helped by it.

\subsubsection{Bi-LSTM Extrapolation}
\label{subsubsec:standard_bilstm_extrap}

Figure~\ref{fig:standard_lstm_extrapolation} reports the Bi-LSTM, trained without the physics-guided modifications, at both
targets: only $R^{2} = 0.81832$ with MAE $= 0.02389$ at $Fo = 0.0009$,
versus a noticeably better $R^{2} = 0.98824$ with MAE $= 0.01050$ at
$Fo = 0.0091$. This gap is the clearest evidence in this study that
backward extrapolation is the harder direction: the same untouched
architecture fits far worse stepping backward than stepping forward
by a comparable distance. Checked against the analytical ground
truth, this is exactly the silent failure the validation methodology
of Section~\ref{subsec:analytical} is designed to catch: a model adequate by
training-window standards whose extrapolation reliability degrades
sharply in one direction.

\begin{figure*}[!t]
\centering
\includegraphics[width=0.85\textwidth]{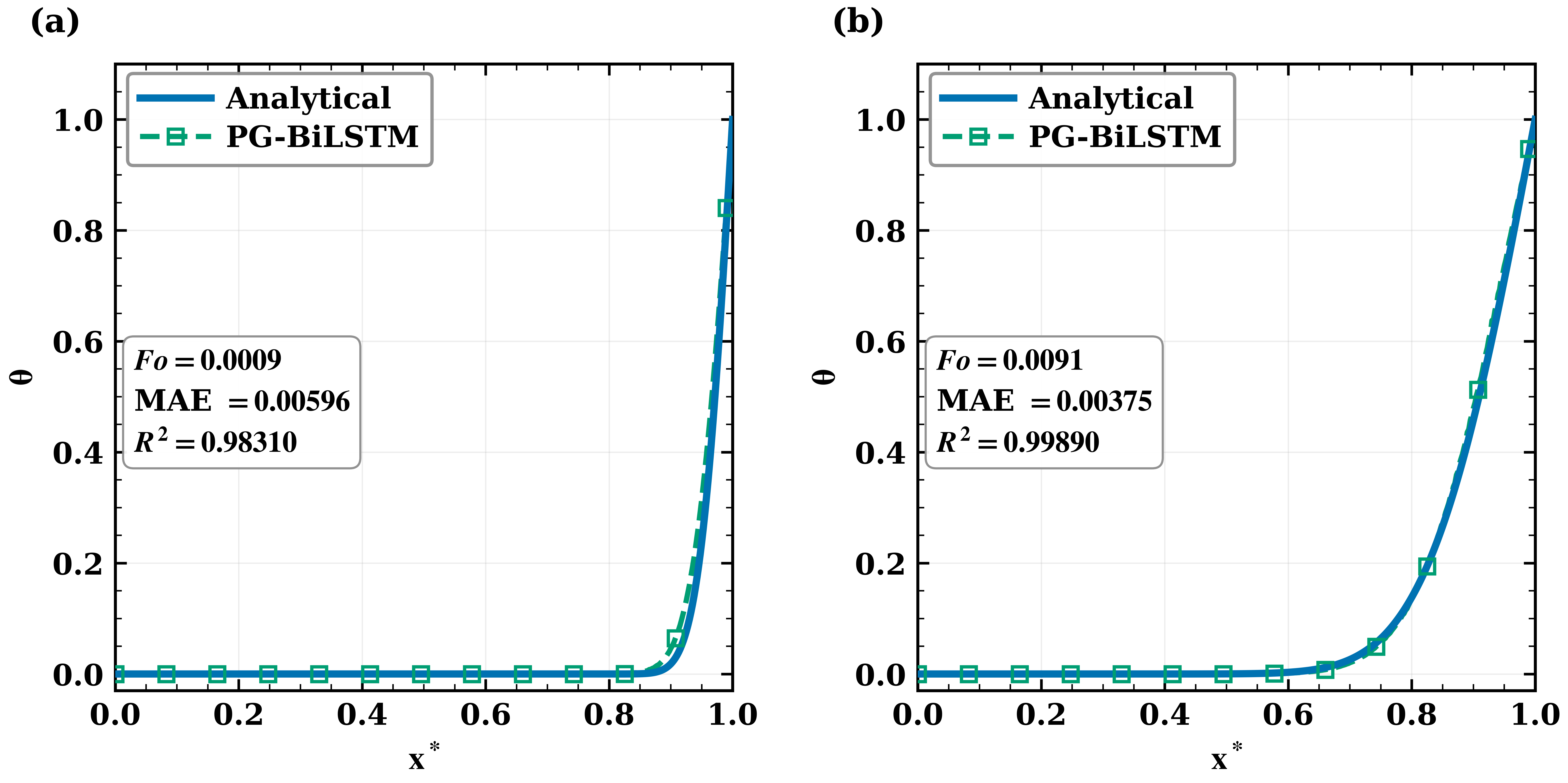}
\caption{Extrapolation from the training range $Fo \in
[0.001,\,0.009]$ using the Physics-guided BiLSTM to (a)~$Fo=0.0009$
and (b)~$Fo=0.0091$, validated against the analytical solution.}
\label{fig:bilstm_target_extrapolation}
\end{figure*}
\begin{figure}
\centering
\includegraphics[width=\linewidth]{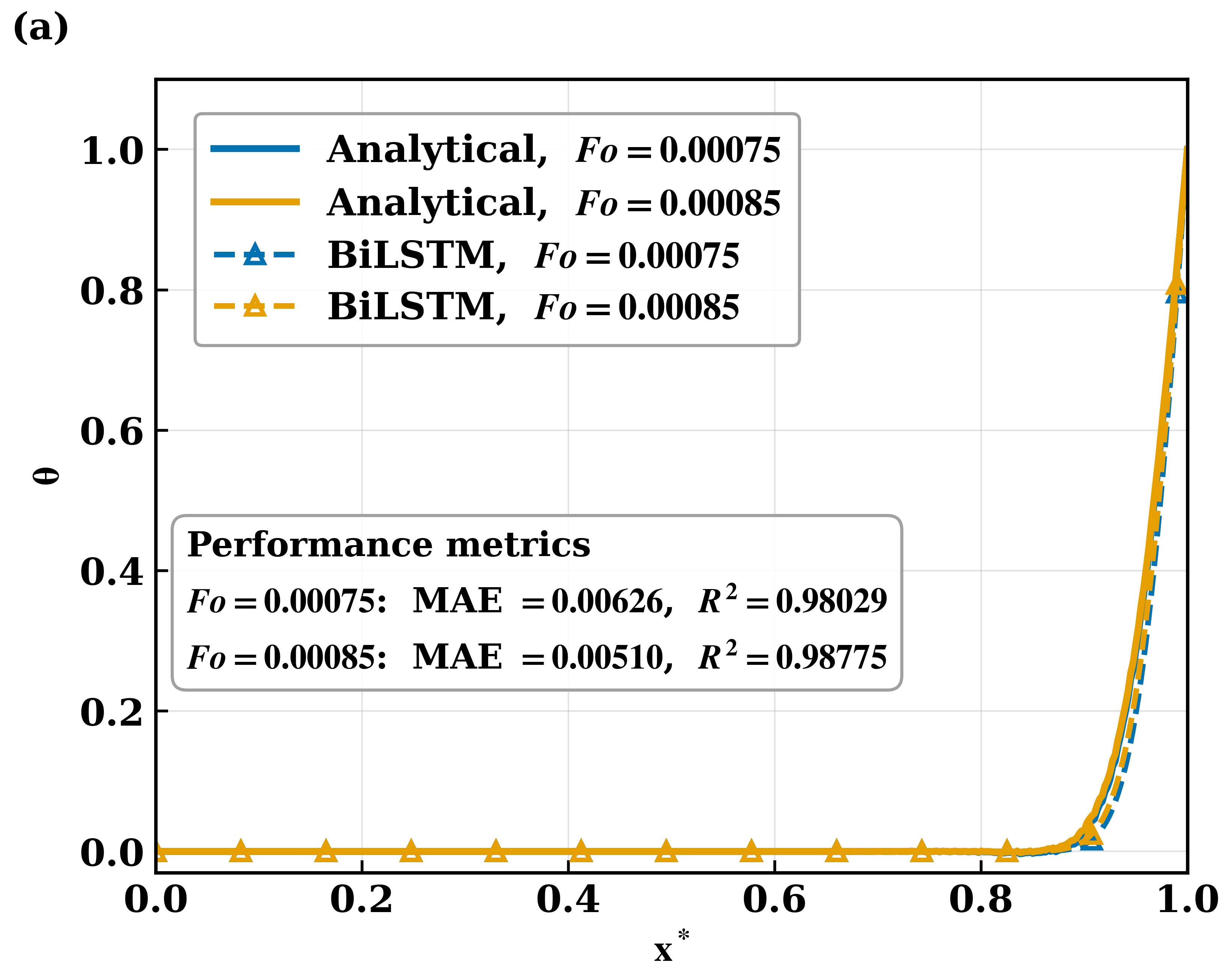}
\caption{Backward extrapolation from the training range $Fo \in
[0.001,\,0.009]$ using the Physics-guided BiLSTM to (a)~$Fo=0.00075$
and (b)~$Fo=0.00085$, validated against the analytical solution.}
\label{fig:bilstm_lower_robustness}
\end{figure}

\subsubsection{PG-BiLSTM Extrapolation at Target Fourier Numbers}
\label{subsubsec:bilstm_targets}

Figure~\ref{fig:bilstm_target_extrapolation} shows the same two
targets after the three physics-guided modifications: $R^{2}$ rises
to $0.98310$ with MAE $= 0.00596$ at $Fo = 0.0009$, a clear
improvement over the standard baseline's $0.81832$, and reaches
$0.99890$ with MAE $= 0.00375$ at $Fo = 0.0091$. The two values are
now much closer together than for the standard model, showing that
the modifications close most of the gap between the easier forward
direction and the harder backward direction rather than helping only
one side. This is the result that matters most for validation: both
extrapolated predictions now pass the same exact-ground-truth check
that the standard model failed at the harder target.

\begin{figure}
\centering
\includegraphics[width=\linewidth]{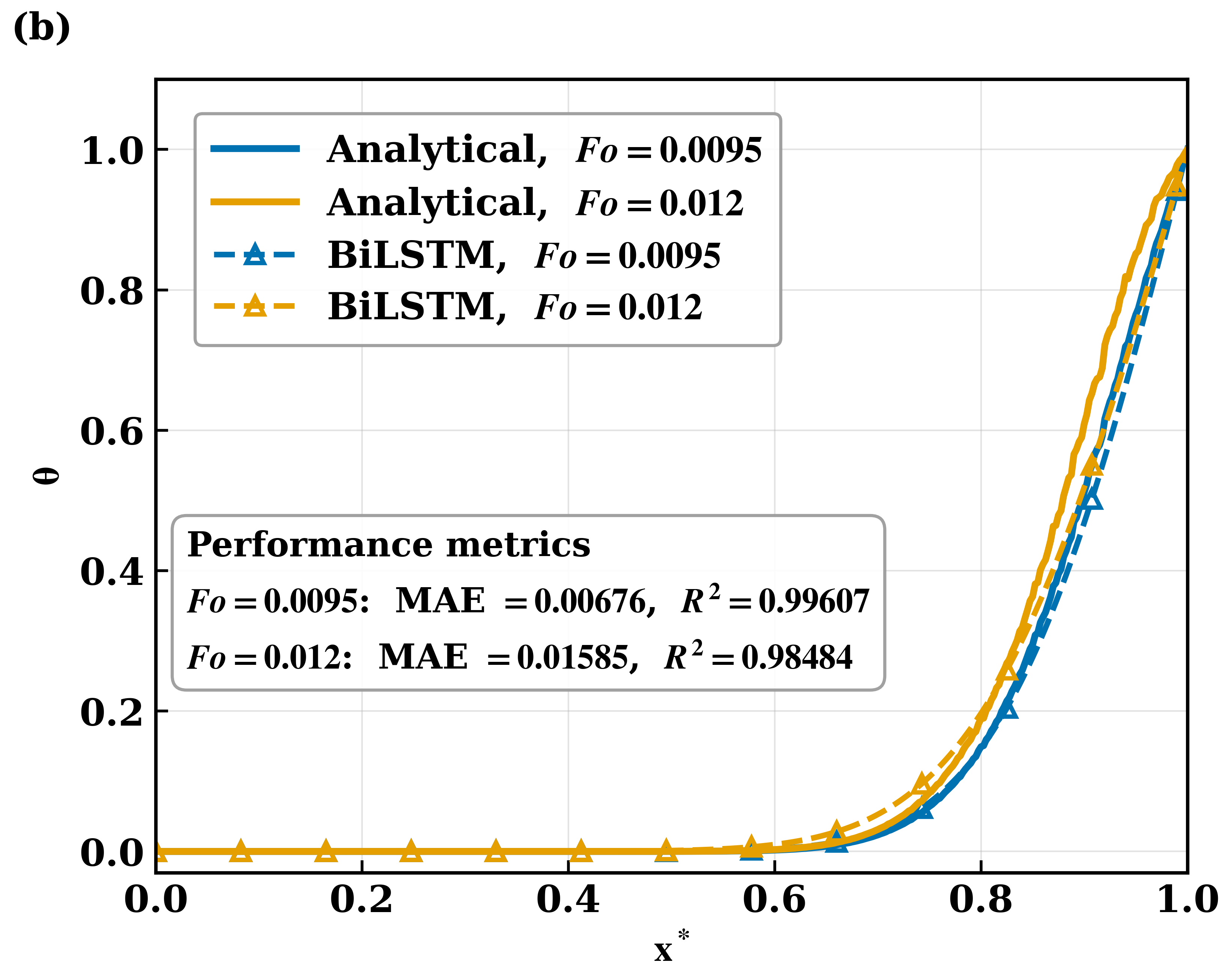}
\caption{Forward extrapolation from the training range $Fo \in
[0.001,\,0.009]$ using the Physics-guided BiLSTM to (a)~$Fo=0.0095$
and (b)~$Fo=0.012$, validated against the analytical solution.}
\label{fig:bilstm_upper_robustness}
\end{figure}
\subsubsection{BiLSTM Extrapolation Across Fourier Numbers}
\label{subsubsec:bilstm_robust}

A single pair of points is not enough to show the model behaves
sensibly as the prediction target moves further from the training
window, so the next two studies extend the test across a range of
Fourier numbers on each side, applying the same progressively
extending validation procedure from Section~\ref{subsec:analytical}.

\paragraph{Lower-regime robustness.}
\label{par:bilstm_lower_robust}

Figure~\ref{fig:bilstm_lower_robustness} reports the Physics-guided BiLSTM across the lower-regime targets: $R^{2} = 0.98029$ with
MAE $= 0.00626$ at $Fo = 0.00075$, and $R^{2} = 0.98775$ with
MAE $= 0.00510$ at $Fo = 0.00085$. The closer target fits slightly
better than the more distant one, matching the expectation that
accuracy should soften gradually rather than collapse suddenly as the
target moves further from the training window, a sign of stable error
propagation through the marching procedure rather than runaway
accumulation.


\begin{figure*}[!t]
\centering
\includegraphics[width=0.85\textwidth]{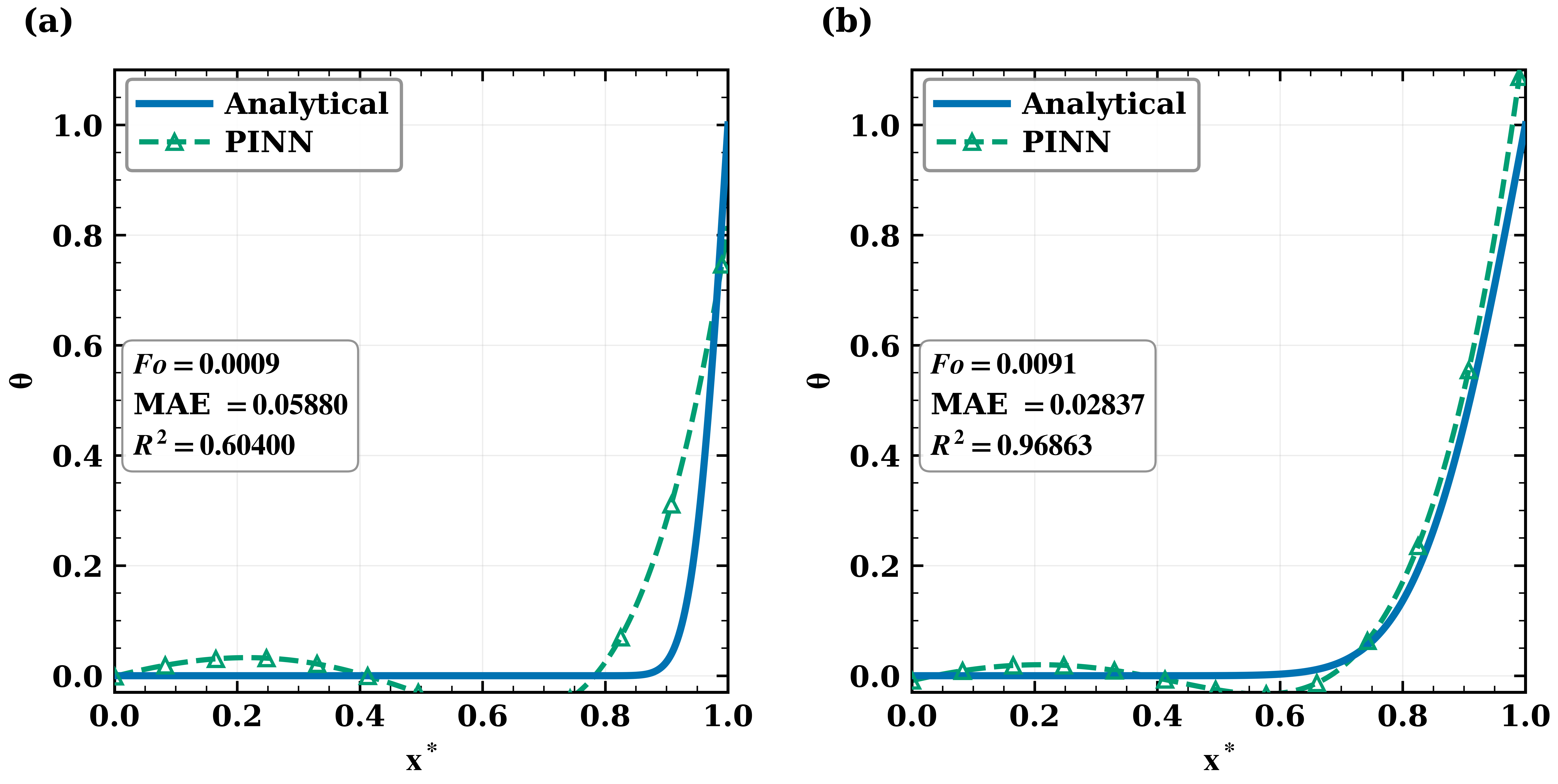}
\caption{Extrapolation from the training range $Fo \in
[0.001,\,0.009]$ using the PINN to (a)~$Fo=0.0009$ and
(b)~$Fo=0.0091$, validated against the analytical solution.}
\label{fig:standard_pinn_extrapolation}
\end{figure*}

\paragraph{Upper-regime robustness.}
\label{par:bilstm_upper_robust}

Figure~\ref{fig:bilstm_upper_robustness} reports the same model across
the upper-regime targets: $R^{2} = 0.99607$ with MAE $= 0.00676$ at
$Fo = 0.0095$, and $R^{2} = 0.98484$ with MAE $= 0.01585$ at
$Fo = 0.012$. Both values remain high, and the accuracy at the
furthest forward target is still comparable to the lower-regime
accuracy in Figure~\ref{fig:bilstm_lower_robustness}, consistent with
forward extrapolation being the gentler direction for this model.
Together, the two sweeps show extrapolation reliability holding up
across a meaningful range on both sides of the training window,
rather than accuracy true only at the two specific points reported
earlier.

\begin{figure*}
\centering
\includegraphics[width=0.85\textwidth]{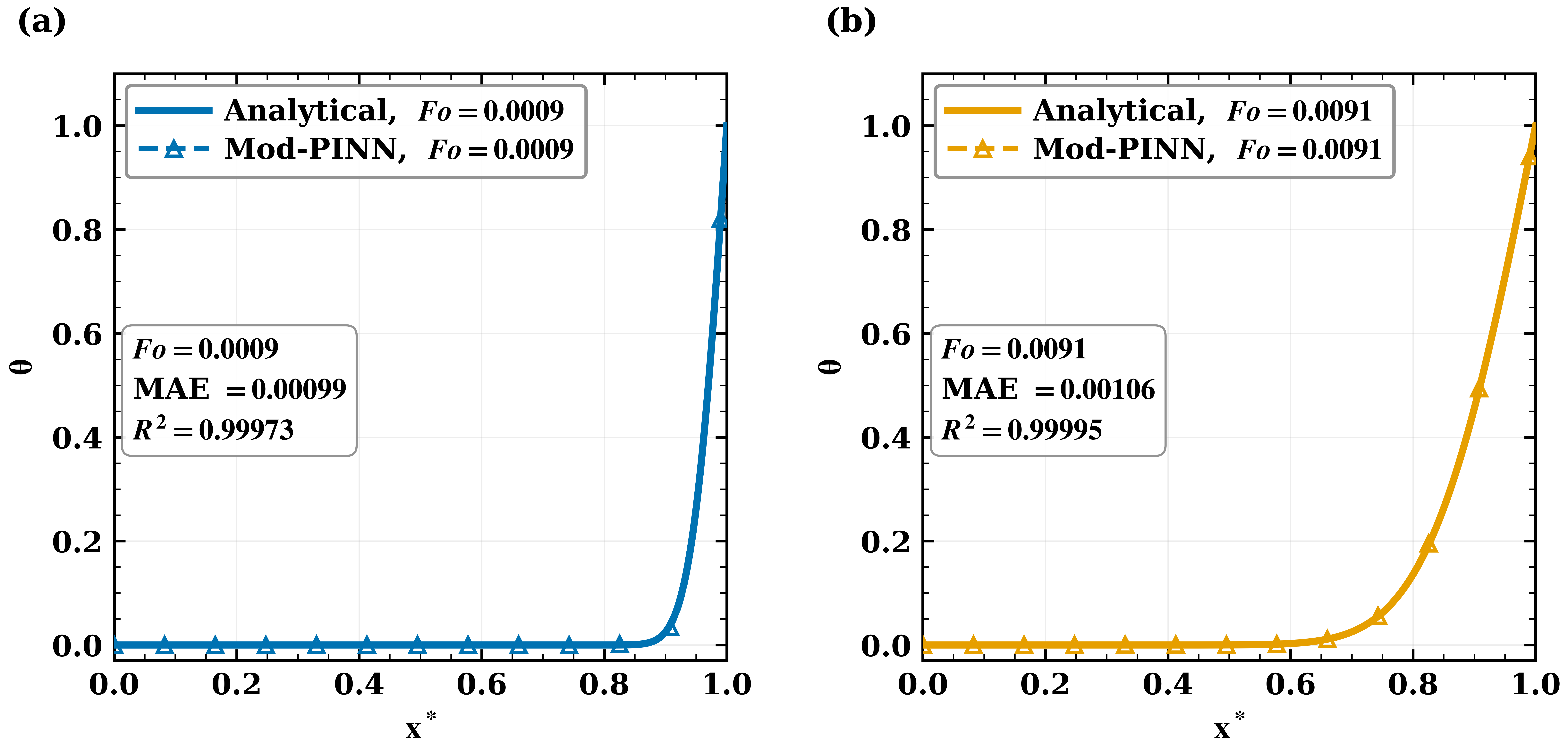}
\caption{Extrapolation from the training range $Fo \in
[0.001,\,0.009]$ using the Modified PINN to (a)~$Fo=0.0009$ and
(b)~$Fo=0.0091$, validated against the analytical solution.}
\label{fig:pinn_target_extrapolation}
\end{figure*}

\subsection{Extrapolation: PINN}
\label{subsec:extrap_pinn}

The PINN is tested through the same lower-target and upper-target
extrapolation pair, followed by the same two robustness sweeps, so
its behaviour can be compared directly against the BiLSTM results
just presented.

\subsubsection{PINN Extrapolation}
\label{subsubsec:standard_pinn_extrap}

Figure~\ref{fig:standard_pinn_extrapolation} reports the PINN, trained without log-time transformation, hard boundary
enforcement, or curriculum learning, at both targets: only
$R^{2} = 0.60400$ with MAE $= 0.05880$ at $Fo = 0.0009$, improving
substantially to $R^{2} = 0.96863$ with MAE $= 0.02837$ at
$Fo = 0.0091$. This contrast is the largest seen anywhere in this
study: the unmodified PINN is by far the weakest of all four
baseline-and-corrected combinations at the lower target, setting up
the clearest case for the three Modified PINN modifications
introduced earlier. Against the analytical ground truth, an $R^{2}$
of $0.60400$ is an extrapolation failure by any reasonable standard,
one a real deployment without an exact reference solution would have
no way of detecting.

\begin{figure*}[!t]
\centering
\includegraphics[width=0.7\textwidth]{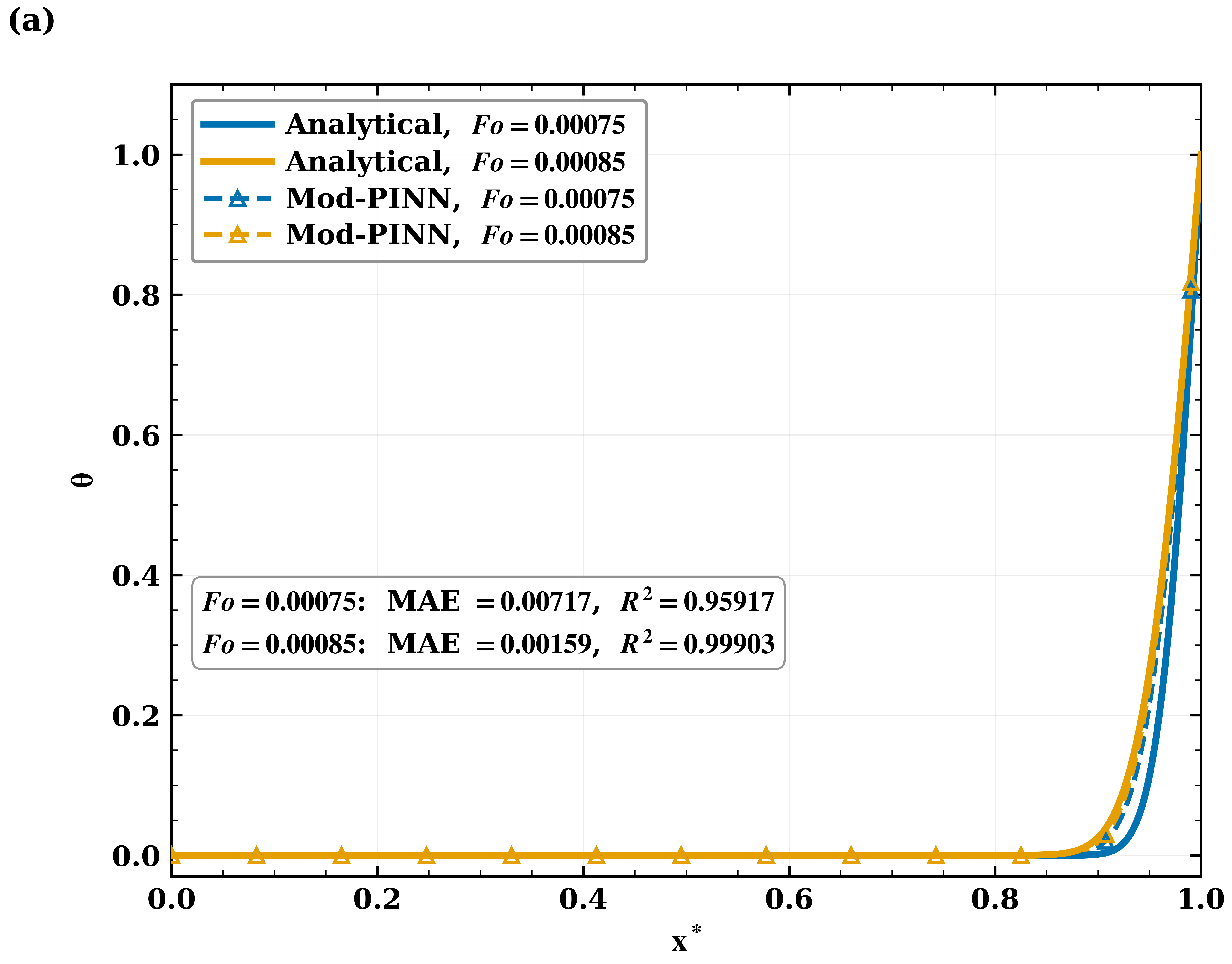}
\caption{Backward extrapolation from the training range $Fo \in
[0.001,\,0.009]$ using the Modified PINN to (a)~$Fo=0.00075$
and (b)~$Fo=0.00085$, validated against the analytical solution.}
\label{fig:pinn_lower_robustness}
\end{figure*}

\subsubsection{PINN Extrapolation at Target Fourier Numbers}
\label{subsubsec:pinn_targets}

Figure~\ref{fig:pinn_target_extrapolation} shows the same two targets
after the physics-guided modifications: $R^{2} = 0.99973$ with
MAE $= 0.00099$ at $Fo = 0.0009$, and $R^{2} = 0.99995$ with
MAE $= 0.00106$ at $Fo = 0.0091$. The lower-target improvement is the
most dramatic single result in this section: $R^{2}$ moves from
$0.60400$ for the PINN to $0.99973$ once the log-time
transformation and hard boundary enforcement are applied, and the two
targets now sit almost on top of each other in accuracy, in sharp
contrast to the PINN's large asymmetry. This is the clearest
demonstration in the paper that an extrapolation failure detected by
validation against ground truth can be traced to a specific
architectural cause and removed by a correspondingly specific fix,
rather than by simply scaling the model up.

\subsubsection{PINN Extrapolation Across Fourier Numbers}
\label{subsubsec:pinn_robust}

\paragraph{Lower-regime robustness.}
\label{par:pinn_lower_robust}

Figure~\ref{fig:pinn_lower_robustness} reports the Modified PINN
across the lower-regime targets: $R^{2} = 0.95917$ with
MAE $= 0.00717$ at $Fo = 0.00075$, and $R^{2} = 0.99903$ with
MAE $= 0.00159$ at $Fo = 0.00085$. As with the BiLSTM lower robustness
study, the target closer to the training boundary fits better,
pointing again to a gradual rather than abrupt loss of accuracy
deeper into the early-time regime. Because the PINN extrapolates
directly rather than marching through its own prior outputs, this
softening reflects the underlying difficulty of the target itself
rather than accumulated step-by-step error, a useful contrast with
the BiLSTM's error-propagation behaviour.

\begin{figure*}[!t]
\centering
\includegraphics[width=0.7\textwidth]{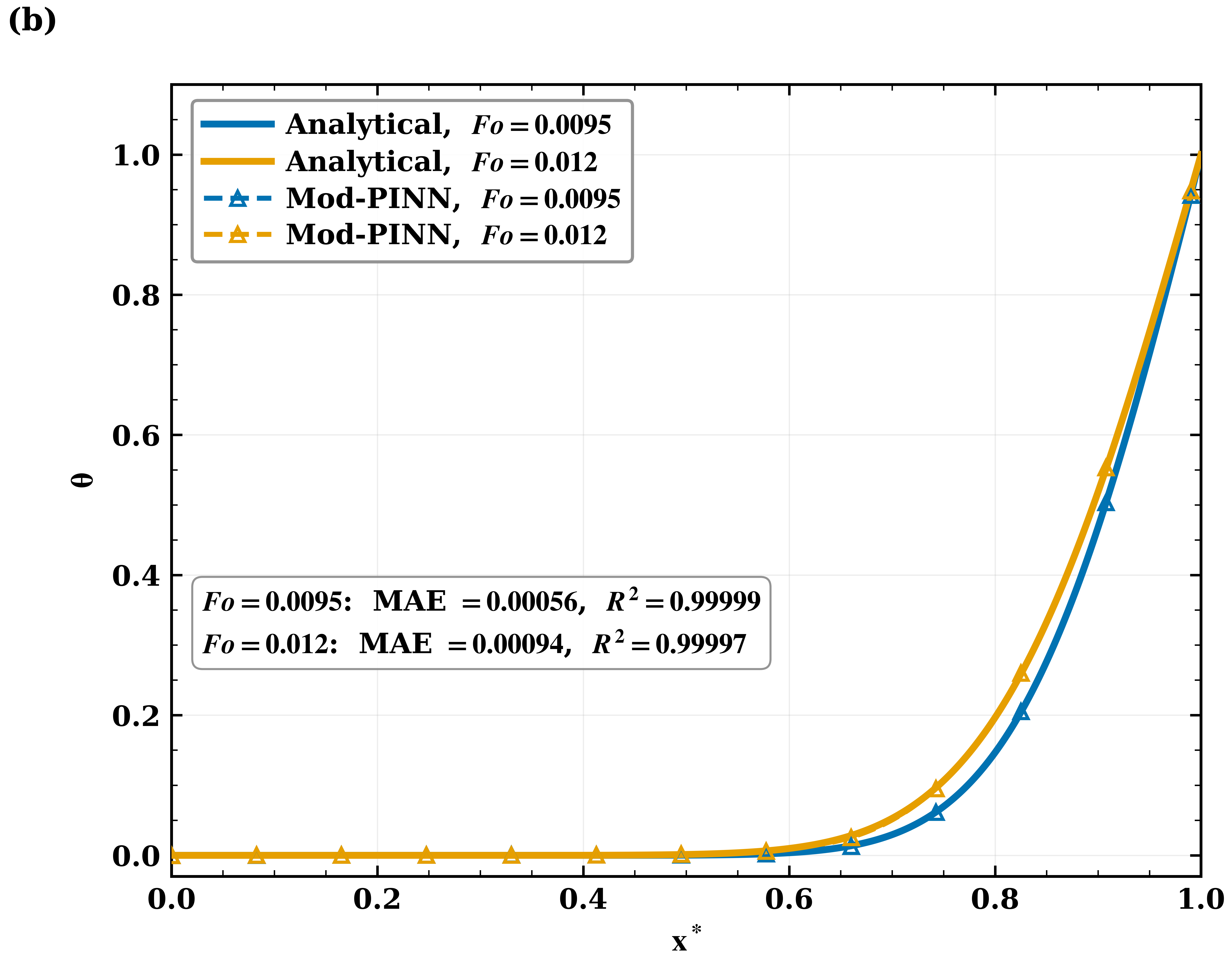}
\caption{Forward extrapolation from the training range $Fo \in
[0.001,\,0.009]$ using the Modified PINN across eight targets
spanning $Fo=0.0091$--$0.0120$, validated against the analytical
solution.}
\label{fig:pinn_upper_robustness}
\end{figure*}
\paragraph{Upper-regime robustness.}
\label{par:pinn_upper_robust}

Figure~\ref{fig:pinn_upper_robustness} reports the Modified PINN
across eight later-time targets spanning $Fo \in [0.00910,\,0.01200]$.
Together with the lower-regime sweep in
Figure~\ref{fig:pinn_lower_robustness} and the target-level results in
Figure~\ref{fig:pinn_target_extrapolation}, this shows the PINN
validated against ground truth across the full range of Fourier
numbers considered in this work, on both sides of the training window,
the breadth of evidence the validation methodology of
Section~\ref{subsec:analytical} was designed to produce.

\begin{figure*}[!t]
\centering
\includegraphics[width=0.68\textwidth]{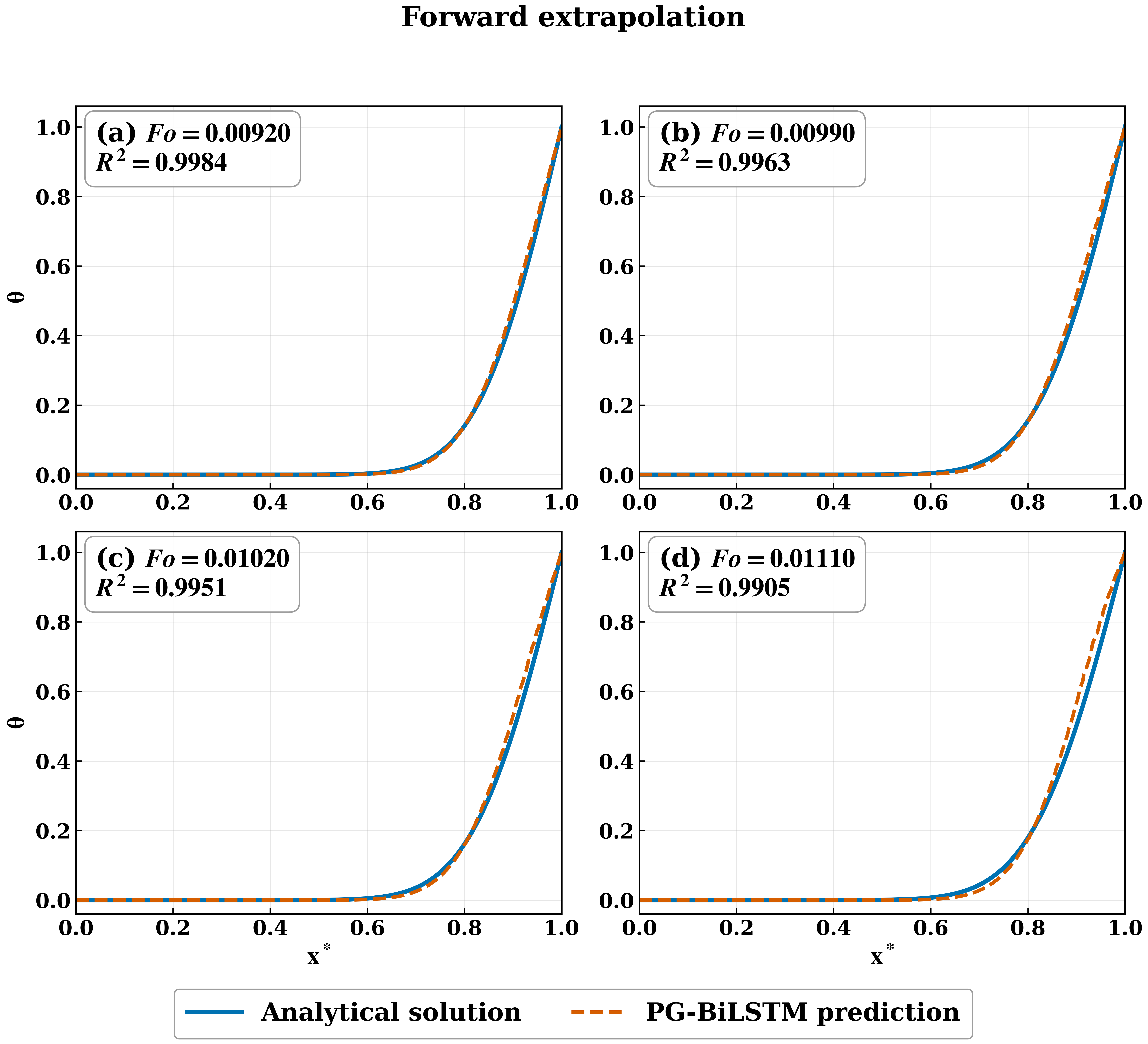}
\vspace{4pt}
\caption{\textbf{} Physics-guided BiLSTM predictions at four increasingly distant forward targets,
(a)~$Fo=0.00920$, (b)~$Fo=0.00990$, (c)~$Fo=0.01020$, and
(d)~$Fo=0.01110$, each checked against the exact analytical solution.}
\label{fig:bilstm_forward_grid}
\end{figure*}
\begin{figure*}
\centering
\includegraphics[width=0.68\textwidth]{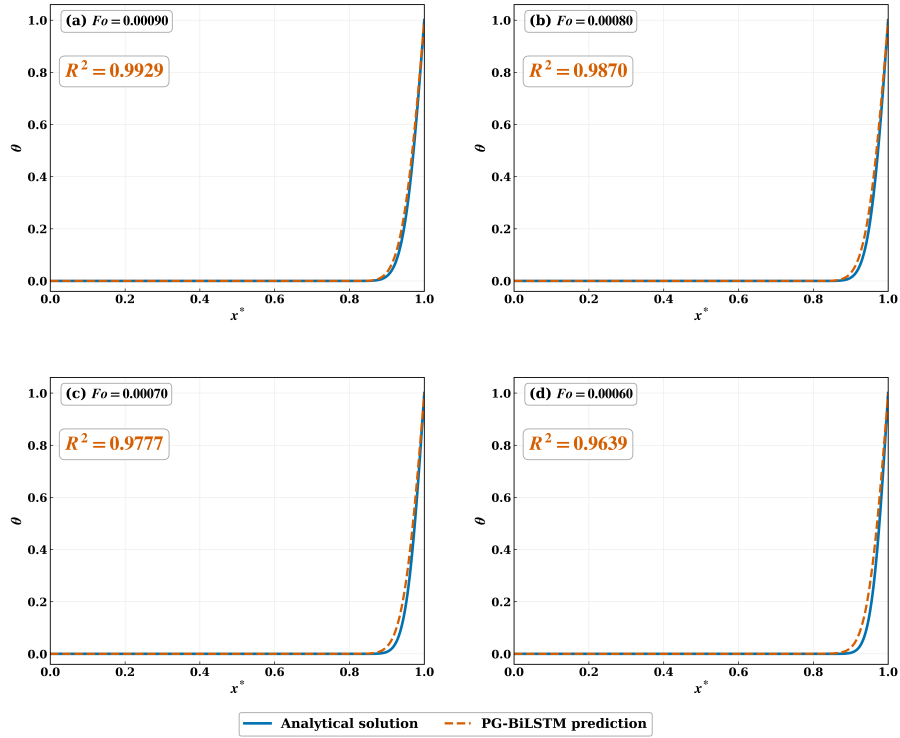}
\vspace{4pt}
\caption{\textbf{} Physics-guided BiLSTM predictions at four increasingly distant backward targets,
(a)~$Fo=0.00090$, (b)~$Fo=0.00080$, (c)~$Fo=0.00070$, and
(d)~$Fo=0.00060$, each checked against the exact analytical solution.}
\label{fig:bilstm_backward_grid}
\end{figure*}

\subsection{Extrapolation Error Growth}
\label{subsec:error_propagation}
The physics-guided BiLSTM offers greater engineering utility than the PINN because it does not require the governing differential equations, which may not always be available. It is therefore important to assess how rapidly extrapolation errors grow.
The lower- and upper-regime sweeps just presented establish that the
Physics-guided BiLSTM remains accurate at several individual targets
on each side of the training window, but a handful of discrete points
still says little about how that accuracy actually evolves as the
prediction horizon is pushed continuously further away. Tracking the model's accuracy as the target Fourier number is advanced, one validated step at a time, progressively farther from the training interval $Fo \in [0.001,0.009]$ in either direction provides a more demanding and informative test. This is the approach adopted in Figures~\ref{fig:bilstm_forward_grid} and~\ref{fig:bilstm_backward_grid}. 

This continuous view allows the asymmetry anticipated in the Introduction (Section~\ref{sec:intro}) to be tested against quantitative evidence. Forward extrapolation was expected to be gentler because diffusion smooths the temperature field and tends to damp small prediction errors. Backward extrapolation was expected to be more challenging because it must reverse this smoothing and reconstruct sharp structures already erased by diffusion. This is a textbook ill-posed reconstruction problem~\cite{Hadamard1902,Beck1985}, in which small uncertainties in the later state can produce large uncertainties in the earlier one.

In the forward direction, Figure~\ref{fig:bilstm_forward_grid} pushes
the target from $Fo=0.009$  to $Fo=0.01110$, roughly
$23\%$ further past the training boundary, and the fit barely moves:
$R^{2}=0.9984$ at panel~(a), eases only to $R^{2}=0.9905$ at panel~(d),
with the predicted curve remaining almost indistinguishable from the
analytical one at every step. This is exactly what
Section~\ref{sec:governing} predicts: as $Fo$ grows, every Fourier
mode in the true solution decays as $e^{-n^2\pi^2 Fo}$, so the physics
itself is becoming simpler at the same time the model is being asked
to march further into it. A small mistake at one step therefore has
little complicated structure left to feed on at the next, and error
grows slowly and stays predictable.

Figure~\ref{fig:bilstm_backward_grid} tells a 
comparable story for backward extrapolation. As $Fo$ shrinks from $0.001$ down to $0.00080$ (which is 20\% beyond the training lower boundary)
$R^{2}$ falls from $0.9929$ to $0.9870$, a slightly larger than the
forward case. The gap between the predicted and analytical curves
near the heated boundary increases as one moves to about 40\% (Fo = 0.0006) outside the training domain. The $R^2$ score is still quite good with a value of 0.9639.
Here, the model must reconstruct a sharper temperature profile from a smoother one. The smoothing caused by diffusion, which makes forward extrapolation easier, must now be reversed. Information that diffusion has erased must be inferred by the model rather than recovered directly from the data. The relaxation-marching scheme   limits error growth by preventing errors from accumulating at every step. However, it cannot make the reconstruction itself easier; it only prevents the errors from growing rapidly as the target approaches the initial singularity at $Fo\to0$.

Taken together, the two sweeps show that extrapolation error grows differently on either side of the training window, reflecting the diffusion physics rather than a peculiarity of the network. Forward predictions remain reliable for longer because diffusion smooths the temperature field, while backward predictions lose accuracy more quickly because the model must reconstruct sharper, more nonlinear, and increasingly boundary-dominated profiles. This confirms the asymmetry anticipated in the Introduction and identifies backward extrapolation as the more demanding test of model reliability. The sequential validation strategy of Section~\ref{subsec}, which checks every step against the exact analytical solution, makes this difference measurable rather than assumed

\subsection{Direct Framework Comparison}
\label{subsec:comparison}

Both models pass the extrapolation-validation test of
Section~\ref{subsec:analytical}, but by different mechanisms, which matters
for which engineering deployment each transfers to. The PG-BiLSTM
steps forward one Fourier number at a time, feeding each prediction
back as input for the next step and checking it against ground truth
at every step, making it reliable for early-time marching where
labelled data already exist nearby. The Modified PINN instead
satisfies the heat equation at collocation points spread across the
domain, predicting at any Fourier number directly without requiring
labelled data at the target, making it the better choice when the
target lies in the later-time region or measured data are scarce.

\begin{table}
\centering
\footnotesize
\setlength{\tabcolsep}{4pt}
\renewcommand{\arraystretch}{1.15}
\caption{Side-by-side comparison of PG-BiLSTM and Modified PINN.}
\label{tab:comparison}
\begin{tabularx}{\columnwidth}{@{}lXX@{}}
\toprule
\textbf{Feature} & \textbf{PG-BiLSTM} & \textbf{Modified PINN} \\
\midrule
Learning mechanism   & Data-driven, sequential          & PDE residual minimisation \\
Boundary conditions  & Learned (boundary-weighted loss) & Enforced exactly by design \\
Extrapolation method & Step-by-step marching            & Direct prediction anywhere \\
Main strength        & Early-time, data-rich regions    & Later-time, data-scarce regions \\
Main challenge       & Error accumulation over steps    & Early-time sharp gradients \\
Training time        & $\approx 45$ min                 & $\approx 120$ min \\
Parameters           & $422{,}801$                      & $18{,}541$ \\
MAE at $Fo=0.0009$   & $0.00596$                        & $0.00099$ \\
MAE at $Fo=0.0091$   & $0.00375$                        & $0.00106$ \\
\bottomrule
\end{tabularx}
\end{table}

Despite having far more parameters ($422{,}801$ versus $18{,}541$),
the BiLSTM trains roughly $2.7\times$ faster ($45$ min versus
$120$ min on the same GPU), since its training steps require only a
forward pass while the PINN must compute PDE residual gradients at
every step. The choice of framework reduces to two questions: how far
is the target Fourier number from the training range, and does
labelled data exist near it? If yes to both, the BiLSTM's validated
step-by-step marching is the more direct route; if not, the PINN's
ability to extrapolate without nearby labelled data is the more
transferable choice.

\subsection{Engineering Applications}
\label{subsec:engineering_applications}

The engineering relevance of these frameworks lies in their ability to predict temperature fields beyond the range covered by available data. In thermal engineering, this is valuable when direct measurements are unavailable or when operating conditions extend beyond those represented in the training set. The following examples therefore illustrate how the frameworks can support prediction in data-limited thermal systems.

The PG-BiLSTM suits inverse heat conduction, where an
earlier temperature history must be reconstructed from later
measurements, and short-horizon thermal control, such as
predicting hotspots in electronics cooling before new sensor data
arrive. Both rely on the same step-by-step marching, validated the
same way against ground truth, shown in
Figures~\ref{fig:bilstm_lower_robustness}
and~\ref{fig:bilstm_upper_robustness}.

The Modified PINN suits settings with sparse or no
instrumentation, such as melt-pool regions in additive manufacturing
or structures under transient aerodynamic or fire loading. Since it
satisfies the governing equation even without labelled data
(Figure~\ref{fig:pinn_upper_robustness}), it can still give a
physically reasonable estimate where the PG-BiLSTM would have no
nearby data to march from. In each setting, the underlying need is
identical to the one this paper is built around: a way to trust a
prediction made outside the range of available data, which is exactly
what the validation methodology developed here is intended to
provide.

Taken together, these results show that both the Physics-guided BiLSTM and the Modified PINN achieve accurate, physically
consistent extrapolation outside the training domain, with every
prediction validated directly against the analytical ground truth
rather than judged by eye. That two architectures built on entirely
different principles, one sequential and data-driven, the other
residual-based and largely data-free, both pass the same validation
test once corrected by the same physical reasoning indicates that the
successful extrapolation demonstrated here is attributable to the
proposed validation framework itself, not to any one network
architecture. This is the central claim the abstract makes, and it is
what these results support.

\section{Conclusions}
\label{sec:conclusions}

This study was not an exercise in predicting transient heat
conduction; the heat-conduction problem served as a controlled
testbed because it offers exact ground truth everywhere, unlimited
data, and physics nonlinear enough to make the test meaningful. This
study established a rigorous methodology for developing and
validating machine-learning models that must extrapolate beyond their
training data, a situation that is unavoidable in engineering
practice and ordinarily impossible to check. By training each model
on a deliberately narrow Fourier-number window and validating every
extrapolated prediction against the analytical solution before
extending the prediction horizon, this work shows that extrapolation
reliability can be certified quantitatively rather than assumed.

Within that methodology, both candidate frameworks needed correction
before their extrapolated predictions could be trusted, and the
specific corrections required reveal why extrapolation fails. A
standard recurrent network struggles because it treats Fourier number
as an ordinary linear input, weighs all spatial locations equally,
and has no safeguard against errors growing during backward marching.
The Physics-guided BiLSTM addresses these issues directly: a
Root-Fourier coordinate aligns the network's notion of time with the
true diffusion timescale, a boundary-weighted loss emphasises the
region where gradients and active modes concentrate, and relaxation
marching bounds error growth. Validated against the analytical
solution at every step, the corrected model extrapolates accurately
well outside its training range, particularly backward, where errors
would otherwise accumulate fastest.

A PINN failed validation for a different reason: the
$Fo^{-3/2}$ singularity makes the PDE residual difficult to optimise
near the initial condition, and soft boundary enforcement lets wall
errors leak into the interior. A log-time transformation removes the
singularity before training begins, and hard boundary-condition
enforcement satisfies the boundary values exactly, freeing the
optimiser to focus on the physics. Once corrected, the PINN's
extrapolated predictions pass the same ground-truth check throughout
the tested range, particularly toward later times where labelled data
are scarce.

The key finding is that the real contribution is not a larger or more
complex network, nor a successful heat-conduction predictor in
itself, but a validated demonstration that extrapolation reliability
can be engineered and checked rather than hoped for. Each
architectural correction traces back to a specific feature of the
governing equation and addresses a specific weakness exposed by the
validation procedure, showing that extrapolation reliability depends
less on model size than on whether the model's structure reflects the
underlying physics, a relationship that can be confirmed
quantitatively whenever an analytically solvable benchmark is
available.

The broader contribution is a transferable framework rather than a
single result on one problem, applicable wherever experiments or
simulations are expensive, available data are limited, extrapolation
beyond that data is unavoidable, and no ground truth exists to check
the result, conditions that describe much of practical engineering.
The methodology calls for proving a model's extrapolation behaviour on
a controlled, analytically or numerically verifiable benchmark first,
using the same sequential train-predict-validate-extend procedure
demonstrated here, before trusting that model where no such
verification is possible, with transient heat diffusion serving only
as the testbed that made this validation possible rather than as the
object of study in its own right.

Future work could extend the same validation methodology, train on a
limited window, check against exact ground truth, and progressively
extend the prediction horizon, to multidimensional heat conduction,
nonlinear thermal systems with temperature-dependent properties, and
other transient physical systems with their own analytically or
numerically certifiable benchmarks. The harder and more important
extension is toward engineering systems where no such exact ground
truth exists, using the confidence built here, on a benchmark
deliberately made more nonlinear and more strictly validated than
most practical datasets, to motivate applying the same physics-guided
philosophy where it cannot be checked exactly.
\section*{CRediT Authorship Contribution Statement}
\setlength{\parindent}{0pt}
\textbf{Ashutosh Yadav:} Methodology, Investigation, Validation,
Data curation, Formal analysis, Software, Writing -- original draft,
Writing -- review \& editing, Visualization.
\textbf{Alok Dubey:} Methodology, Writing -- review \& editing. 
\textbf{Prodyut Ranjan Chakraborty:} Conceptualization, Resources,
Writing -- review \& editing, Supervision, Project administration.
\textbf{Harshal Deepak Akolekar:} Conceptualization, Software, Formal Analysis, Resources
Writing -- review \& editing, Supervision.

\printcredits

\section*{Declaration of Competing Interest}
\setlength{\parindent}{0pt}
The authors declare no known competing financial interests or personal
relationships that could have appeared to influence the work reported
in this paper.

\section*{Declaration of Generative AI and AI-Assisted Technologies}
The authors declare that generative AI and AI-assisted technologies were used to improve the grammar and readability of this manuscript. All scientific content remains the responsibility of the authors.

\section*{Acknowledgements}
\setlength{\parindent}{0pt}
The authors thank the Indian Institute of Technology Jodhpur for
providing computational resources, including access to an NVIDIA
Tesla T4 GPU through the Google Colab Pro environment. A.Y.\
acknowledges support through the IIT Jodhpur M.Tech programme in
Mechanical Engineering.

\section*{Data Availability}
\setlength{\parindent}{0pt}
The analytical benchmark data are reproducible from the Fourier
series solutions in Section~\ref{sec:governing} and are available
from the corresponding author upon reasonable request.

\end{document}